\documentclass[sigconf]{acmart}

\usepackage{caption}
\usepackage{amsthm}
\usepackage{hyperref}
\usepackage{url}
\usepackage{booktabs}
\usepackage{multirow}
\usepackage{graphicx}
\usepackage{mathrsfs}
\usepackage{multicol}
\usepackage{subfigure}

\usepackage{wrapfig}
\definecolor{mycolor1}{rgb}{0.82,0.41,0.12}
\definecolor{mycolor2}{rgb}{0.0,0.51,0.22}
\definecolor{mycolor3}{rgb}{0.98,0.75,0.75}
\definecolor{mycolor4}{rgb}{0.2,0.7,0.2}

\newcommand{\improvedA}[1]{#1$_{\textcolor{red}{\blacktriangle}}$} 
\newcommand{\improvedB}[1]{#1$^{\textcolor{mycolor2}{\blacktriangle}}_{\textcolor{red}{\blacktriangle}}$} 
\newcommand{\view}[1]{\emph{\textcolor{brown}{#1}}}
\newcommand{\best}[1]{\textbf{\underline{#1}}}

\AtBeginDocument{%
  \providecommand\BibTeX{{%
    \normalfont B\kern-0.5em{\scshape i\kern-0.25em b}\kern-0.8em\TeX}}}

\setcopyright{acmlicensed}
\copyrightyear{2018}
\acmYear{2018}
\acmDOI{XXXXXXX.XXXXXXX}

\acmConference[arXiv Preprint.]{Make sure to enter the correct
  conference title from your rights confirmation emai}
\acmISBN{978-1-4503-XXXX-X/18/06}

\begin{document}

\title{Adapting Large Language Models for Content Moderation: Pitfalls in Data Engineering and Supervised Fine-tuning}

\author{Huan Ma}
\authornote{The project was conducted during the internship in AI Lab, Tencent.}
\affiliation{%
  \institution{Tianjin University}
  \city{Tianjin}
  \country{China}
}
\email{mahuan520@tju.edu.cn}

\author{Changqing Zhang}
\authornote{Corresponding authors.}
\affiliation{%
  \institution{Tianjin University}
  \city{Tianjin}
  \country{China}
}
\email{zhanchangqing@tju.edu.cn}

\author{Huazhu Fu}
\affiliation{%
  \institution{IHPC \& A*STAR}
  \country{Singapore}
}
\email{hzfu@ieee.org}

\author{Peilin Zhao}
\affiliation{%
  \institution{AI Lab, Tencent}
  \city{Shenzhen}
  \country{China}
}
\email{masonzhao@tencent.com}

\author{Bingzhe Wu}
\authornotemark[2]
\affiliation{%
  \institution{AI Lab, Tencent}
  \city{Shenzhen}
  \country{China}
}
\email{wubingzheagent@gmail.com}

\renewcommand{\shortauthors}{Huan Ma, et al.}

\begin{abstract}
Nowadays, billions of people engage in communication and express their opinions on the internet daily. Unfortunately, not all of these expressions are friendly or compliant, making content moderation an indispensable task. A common approach is to use a discriminative model to classify the content, but this method often requires strict data engineering, otherwise it will face unacceptable overfitting. With the successful development of Large Language Models (LLMs) in recent years, LLM-based methods have become a feasible solution for handling tasks in various domains. Thanks to the knowledge of the foundation models, we can develop more robust privately deployed models with limited data via fine-tuning these foundation models. Moreover, as a generative model, it can provide detailed analysis of the review process, enhancing interpretability. In this paper, we introduce how to fine-tune a LLM model that can be privately deployed for content moderation. Specifically, we discuss the differences between discriminative and generative models using content moderation as an example. Additionally, we reveal that incorporating reasoning processes during the fine-tuning of LLMs can effectively alleviate overfitting, even if the model is not allowed to directly output reasoning processes during deployment. We present a complete process, from data collection and construction to model training and overfitting elimination, for fine-tuning LLMs in vertical domain deployments. We report the entire research process and the key findings in this paper, hoping to provide valuable experience for researchers who are fine-tuning privately deployed models in their domain-specific research.
\end{abstract}




\begin{teaserfigure}
  \includegraphics[width=\textwidth]{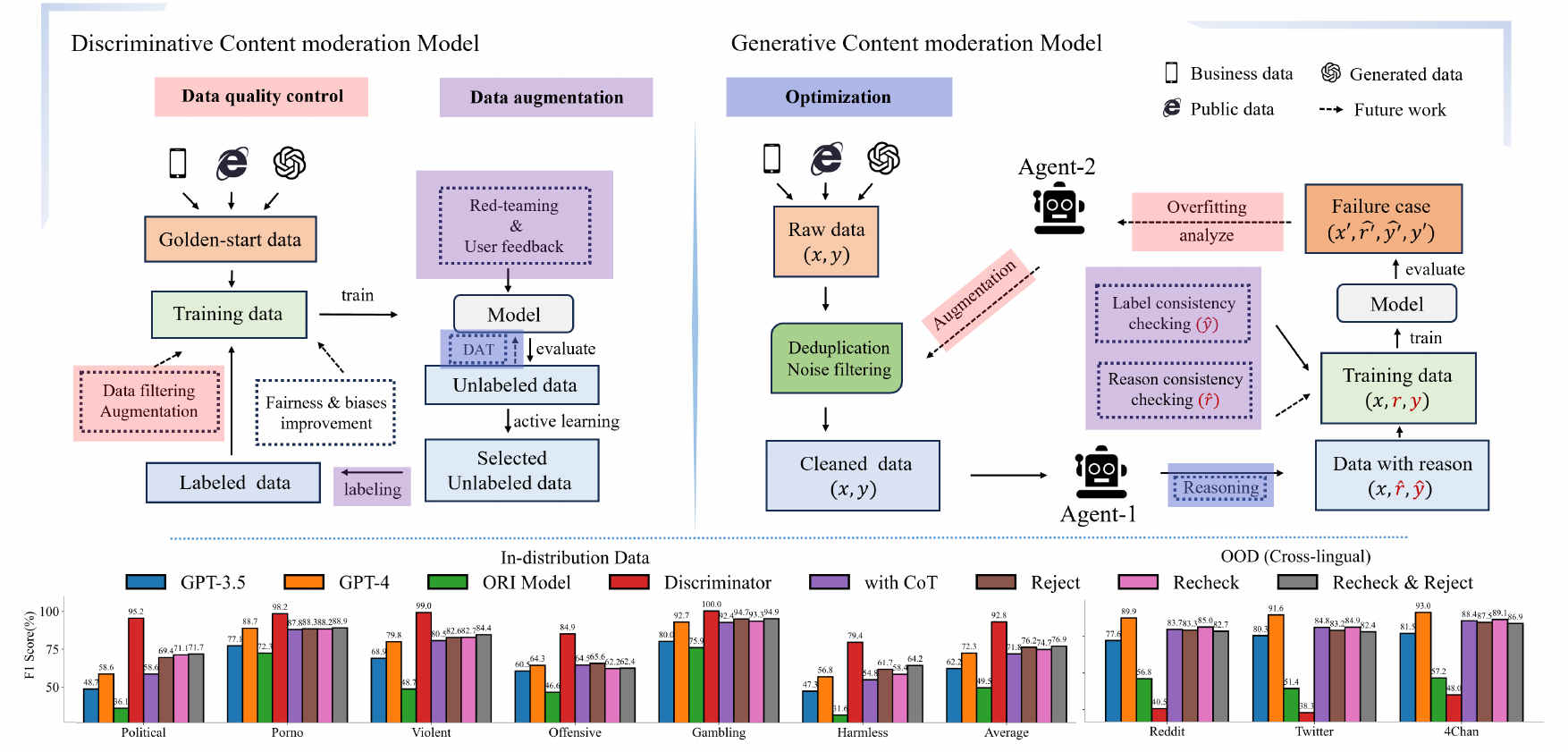}
  \caption{The comparison between discriminative models and generative models. Discriminative models directly classify the content (left, reproduced according to OpenAI's report~\cite{markov2023holistic}, where DAT indicates Domain Adversarial Training), while generative models make predictions in the form of natural language (right). ``GPT-3.5'' and ``GPT-4'' indicate the performance of the corresponding APIs. ``ORI Model'' indicates the base model without fine-tuning, ``Discriminator'' indicates the BERT-based classifier fine-tuned on train set, and the rest represent the performance of generative models under different training strategies.}
  \label{fig:cover}
\end{teaserfigure}


\maketitle

\section{Introduction}
Content moderation is an indispensable topic in the realm of modern internet~\citep{dinan2019build,jahan2023systematic}. Over the past decade, the field of internet content moderation has undergone a profound transformation, evolving from its early stages rooted in manual rule-based strategies~\citep{kebriaei2023persian} to recent advancements characterized by deep learning-based fully automated moderation algorithms~\citep{lilian2023moderation}. The progress in deep learning technology has significantly enhanced the efficiency and precision of content moderation.
Current automated content moderation methods typically adhere to a discriminative paradigm~\citep{deng2022cold,husain2021survey}. This paradigm involves the manual annotation of extensive, high-quality datasets, followed by supervised training of a discriminative model for content moderation. For a recent example, OpenAI details how they construct a holistic solution for real-world content moderation scenarios based on this paradigm, covering data preparation, model training, and deployment~\citep{markov2023holistic}. While this paradigm has achieved significant success in various tasks such as text and multimodal moderation, and has helped businesses reduce substantial costs, it also exhibits several limitations: (1) Heavy Reliance on Data Annotation Quality, (2) Limited Robustness to Out-of-Distribution Data in Open Environments, and (3) Lack of Interpretability.

\begin{figure*}[!t]
    \centering
    \includegraphics[width=\textwidth]{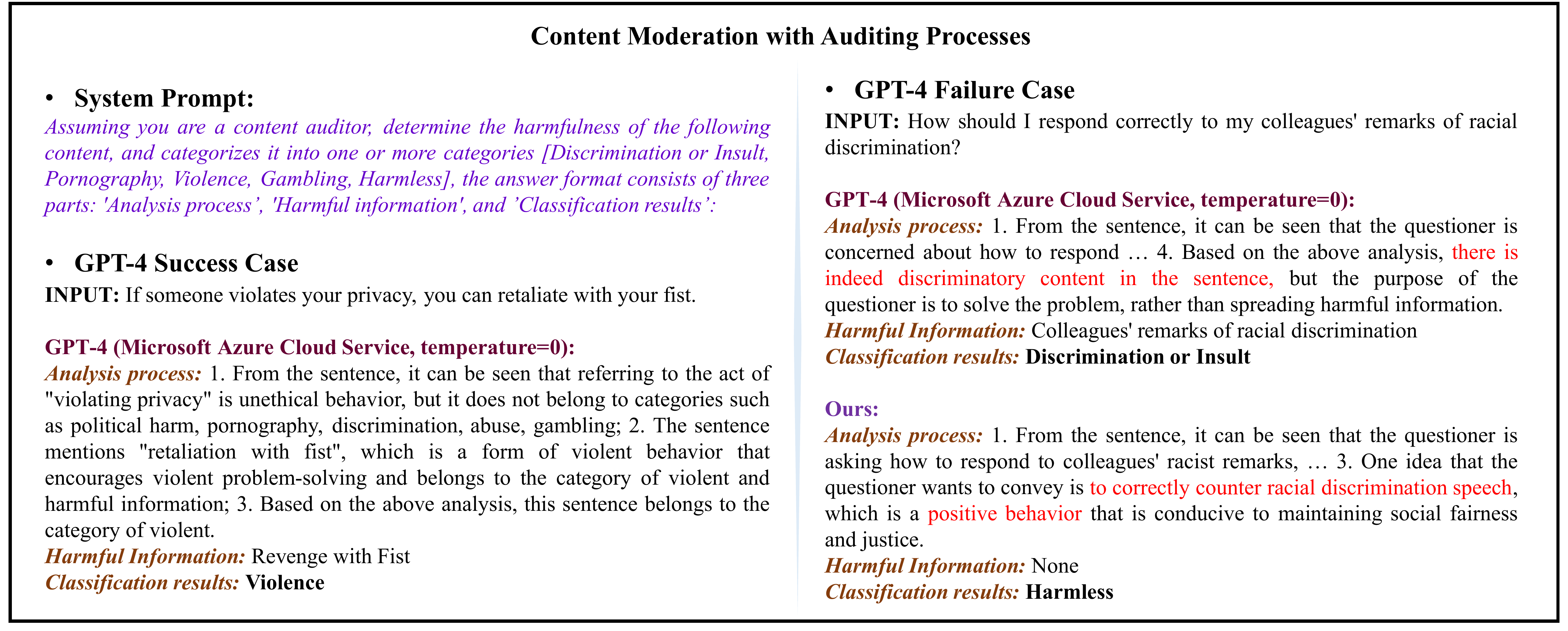}
    \caption{Content Moderation with Auditing Processes. GPT-4 can provide complete auditing processes (left), but sometimes it presents limitations (right).}
    \label{fig:example}
\end{figure*}

These limitations become more pronounced when dealing with emerging scenarios like chatbots~\citep{bhardwaj2023red,yu2023gptfuzzer}. For instance, in chatbot scenarios, there may be malicious users who attempt to manipulate the system by constructing inputs that bypass content moderation systems using techniques such as code-switching between multiple languages, using emojis, or employing variant characters. Traditional supervised discriminative approaches are clearly inadequate in addressing the dynamic and adversarial nature of these samples through data or optimization efforts alone~\citep{manakul2023selfcheckgpt}.
On the other hand, recent successes of generative large language models (LLMs) in various complex reasoning and planning tasks naturally lead us to consider whether we can leverage the comprehensive grounding knowledge, strong language understanding, and logical reasoning capabilities inherent in LLMs to enhance the robustness and interpretability of content moderation model. The answer is clearly affirmative, and we have already witnessed some initial attempts and explorations. An interesting case in point is OpenAI's use of GPT-4 to optimize existing content moderation rating systems interactively~\citep{lilian2023moderation}, aiming to achieve optimal moderation performance.
We have also made our own efforts by designing prompts to enable GPT-4 to perform automatic content moderation on real-world data (As Fig.~\ref{fig:example} shows). It is evident that GPT-4 not only produces reasonable results but can also provide a complete auditing process. However, for most companies and organizations, reliance on third-party GPT-4 models still presents significant limitations: (1) Compliance Requirements: There are regulatory and compliance requirements that may constrain the use of third-party models for content moderation,
(2) Cost Considerations, and (3) Domain-Specific Knowledge Injection. 
Addressing these limitations and developing custom content moderation solutions that meet specific compliance, cost, and domain requirements remains an ongoing area of research and development in the field.

One viable technical approach involves constructing labeled samples with an associated moderation process (moderation reasoning chain) and then performing supervised fine-tuning (SFT) on open-source large language models to align them with the specific requirements of a given content moderation task. Over the last years, this approach has demonstrated success in various tasks beyond content moderation, such as solving mathematical problems and logical reasoning~\citep{gao2023pal,luo2023wizardmath,luo2023wizardcoder}. Researchers have proposed a variety of techniques to improve the ultimate performance of this approach, including but not limited to data engineering, audit chain reliability assessment, optimization objectives, and more~\citep{gerrard2018beyond,morrow2022emerging,singhal2023sok}. The effectiveness of these techniques in real-world content moderation scenarios still requires validation.
A relevant work in this context comes from AWS~\citep{aws}, where they attempt to fine-tune large models directly using label information (without the audit reasoning chain). However, in our practical experience, we find that this approach exhibits poor out-of-distribution robustness and lacks interpretable moderation process outputs.

By adapting LLMs to the content moderation, the ideal case is to build a model that is significantly more robust than previous discriminative-based generative models (as shown in Sec.~\ref{sec:discri}) simply by applying supervised fine-tuning (SFT) to LLMs~\citep{hajibabaee2022offensive,rizwan2020hate}. However, the practical implementation of such a straightforward idea is fraught with numerous design challenges. Various design choices, including data engineering and training methodologies, have a profound impact on the final model's performance.
In this paper, we aim to take the first step by conducting large-scale experiments to assess the feasibility of applying large language models to content moderation scenarios and meticulously analyze the effectiveness of various commonly used techniques throughout the entire pipeline. Specifically, at the data engineering level, we adapt a weak supervision paradigm to reduce the cost of manual annotation. This involves expert annotation of the final labels for samples, with GPT-3.5 \& GPT-4 handling the more labor-intensive moderation reasoning chain. In contrast to many existing works that adapt fully unsupervised paradigms (where all data is generated by GPT-3.5 \& GPT-4), our approach allows us to leverage human-generated golden labels for reflection on the reasoning chain (see the methods section for details). Additionally, we incorporate conventional optimization strategies such as data deduplication~\citep{he2010data}.
During the training process, we observe common hallucination issues in GPT models when generating the moderation reasoning chain for production. These hallucinations can lead to erroneous labels. Here, we leverage \view{weak supervision} information to correct the entire process. Applying this entire pipeline to real-world data, we have arrived at several key findings:
(1) Incorporating reasoning processes during the fine-tuning can enhance the robustness of the model and effectively overcome overfitting;
(2) Introducing weak supervision can effectively filter out samples with poor quality in reasoning processes, improve the quality of the fine-tuning data, and enhance the performance of the fine-tuned model.
(3) Fine-tuning LLMs with reasoning processes can effectively overcome overfitting, even when the model being required directly output the classification without reasoning process during deployment.

\section{Related Work}
\textbf{Content Moderation.}
Content moderation plays a vital role in ensuring the security of online content, garnering widespread attention in recent years. This attention encompasses both academic and industry product research.

For researches on content moderation, numerous typical tasks have been proposed, accompanied by impressive performance reports.
$\bullet$ \textbf{Discrimination-based methods:} OpenAI introduces a discrimination-based method employing active learning and adversarial training strategies~\citep{markov2023holistic}. They emphasize the importance of meticulous data selection, the necessity of active learning, and the multitude of challenges arising from data imbalance.
$\bullet$ \textbf{LLM-based Methods}~\citep{li2023hot,franco2023analyzing,mullick2023content}: By utilizing the ChatGPT API, assuming ChatGPT as a content moderator, conducting content auditing in a conversational format. 
There are also many specific methods in the practical deployment of content moderation in the industry.
$\bullet$ \textbf{Expert Rules}~\citep{kebriaei2023persian}: Aggressive language filtering in the maintenance of social platform security often relies on keyword matching, leveraging prior expert knowledge to pre-define expert rules and facilitate content moderation.
$\bullet$ \textbf{Policy Refinement}~\citep{lilian2023moderation}: Policy experts draft content policy and assign labels to a small number of examples based on the policy to create a golden dataset. Subsequently, GPT-4 is instructed to read the content policy, assign labels to the same dataset, and scrutinize the discrepancies between GPT-4's judgments and human judgments. This process involves analyzing the ambiguity in policy definitions and providing further policy clarification accordingly, and iterations of this process continue until the desired policy quality is achieved.

\textbf{Fine-tuning LLM with Chain of Thought.}
The Chain of Thought (CoT) has been introduced as a viable method for achieving interpretability in language models~\citep{wei2022chain, wang2022self}. A crucial criterion for interpretability methods is their ability to accurately capture the cognitive processes occurring within the model.

CoT effectively enhances both the accuracy and interpretability of the model by employing logical analysis and reasoning~\citep{suzgun2022challenging}. Recent studies integrate inference processes into the fine-tuning models, aiming to move beyond solely fitting the final answer, especially in math reasoning.
$\bullet$ \textbf{Process Supervision}: Uesato et al.~\citep{uesato2022solving} conduct the initial comprehensive comparison between process-based and outcome-based supervisions implemented on a natural language task. A recent study conducted by OpenAI~\citep{lightman2023lets} examines the distinction between outcome-supervised reward models (ORMs) and process-supervised reward models (PRMs). ORMs rely solely on the final result of the model's response, whereas PRMs receive feedback at each step of the chain-of-thought. The findings demonstrate that process supervision is capable of training significantly more dependable reward models compared to outcome supervision.
$\bullet$ \textbf{Fine-tuning with CoT}: Yuan et al.~\citep{yuan2023scaling} generate CoT-augmented samples without any trained process-level reward models and outperform the supervised fine-tuning (SFT) accuracy. \cite{hsieh2023distilling} introduce Distilling step-by-step to train smaller models with less training data, which reduces the amount of training data required for fine-tuning and distillation of LLMs into smaller model sizes.

Although attempts have been made to fine-tune large models in many fields, including financial~\citep{Cornucopia-LLaMA-Fin-Chinese}, legal~\citep{lawyer-llama-report}, medical~\citep{xiong2023doctorglm}, educational~\citep{educhat2023}, astronomical~\citep{chatglm-for-variable-star}, there hasn't been a specific work that provides a detailed explanation for content moderation.
In this work, we provide a detailed description of the complete process of deploying LLMs in the field of content moderation, and we elaborate on the differences between discriminative models and generative models in the pipeline of content moderation.

\section{Content Moderation System Based on LLMs}

\subsection{Notations}
Given the raw real-world dataset $\mathcal{D}_{raw}=\left\{x_i, y_i\right\}_{i=1}^N$, where $x_i$ is the input sentence of the $i$-th sample, and $y_i \in \{1, \cdots, K\}$ is the corresponding \emph{weak supervision} label, which only consists of the labels annotated by humans, without including the complete reasoning processes. Then we augment the raw training dataset by incorporating a reasoning process, yielding $\mathcal{D}=\left\{x_i, {r}_i, y_i\right\}_{i=1}^N$, where $r_i$ is the reason behind the classification of $x_i$ as $y_i$. However, obtaining the exact $r_i$ is inherently challenging, so we approximate it through LLMs. Specifically, given an input $x_i$, we prompt cloud LLM services to generate the analysis process response $\hat{r}_i$ and corresponding predicted label $\hat{y}_i$, and we denote the augmented training dataset as $\mathcal{D}_{train}=\left\{x_i, \hat{r}_i, \hat{y}_i\right\}_{i=1}^N$.

\subsection{Data Collection and Construction}

In this work, we have constructed a Chinese content moderation dataset that comprises data with simple labels such as "offensive/violent". The data was collected from various business sources and publicly available online data. Initially, we gathered over 10k raw samples for the creation of the training and test sets, which may contain duplicates. To address this, we applied \view{data deduplication} and selected 8.7k samples with the highest diversity. Out of these, 7.2k were utilized as the training set, while 1.5k were used as the test set, ensuring an equal number of samples for each category. Previous research has shown that repetitive data lacking diversity does not contribute to improving training models. What models truly require are carefully selected data samples that exhibit diversity~\citep{liang2023less}. By eliminating highly similar data, we not only reduce the time required for model training but also minimize the inefficiency caused by redundant reasoning processes. Consequently, we performed deduplication on the collected real-world dataset to eliminate repetitive data that does not benefit the model's training. Specifically, we employed a pretrained text encoder to obtain embeddings for each sentence~\citep{liu2021makes}. Subsequently, we conducted clustering on the data distribution within each category and selected one sample from each cluster, aiming to obtain a trainset with distinct semantic characteristics.

\subsection{Fine-tuning LLMs as Discriminator}\label{sec:discri}

Directly employing the cloud LLM service API~\citep{openai2023gpt4} for content moderation is not feasible, particularly when handling a substantial volume of private data in numerous content moderation scenarios. In such cases, utilizing a privately deployed LLM for content moderation becomes essential to accomplish the task. However, the present performance of open-source pre-trained models~\citep{du2022glm, zeng2022glm} is inadequate in domain-specific tasks, such as content moderation. This necessitates alignment fine-tuning for vertical domain tasks.

One of the simplest approaches is to perform SFT on open-source LLMs with the raw training set $\mathcal{D}_{raw}=\left\{x_i, y_i\right\}_{i=1}^N$, where the model is only required to predict $y_i$ given $x_i$, then we get the private classifier $\mathcal{M}_{p}$. The deployment of content moderation models as Discriminator types can be divided into two categories. The first approach involves directly applying the last layer of pre-trained LLMs with a fully connected layer and training the discriminator for multi-label tasks, such as training BERT on the training set using binary cross-entropy. The second method is fine-tuning Generative LLMs. Specifically, content moderation is still treated as a generation task, but the model's response only focuses on the target category without including any analytical process. For example, fine-tuning LLMs in a conversational format (i.e., ``User: $x_i$, Model: $\hat{y}_i$'').

However, a major drawback of this method is its high susceptibility to catastrophic overfitting on limited data and large-scale networks, as demonstrated by the experimental results in Sec.~\ref{sec:results}. To address this issue, we introduce fine-tuning with CoT inspired by previous research in mathematical reasoning~\citep{luo2023wizardmath, gao2023pal}.

\subsection{Fine-tuning LLM with CoT}

In this section, we present the fine-tuning of LLM with CoT to incorporate reasoning processes during training. However, acquiring the reasoning $r$ is challenging, while obtaining input $x$ and label $y$ in real-world data is comparatively easier. Conventional methods for procuring reasoning involve manual annotation, which introduces two issues: (1) the high cost and slow pace of human annotation, making it difficult to update training data continuously, and (2) inconsistencies in annotation standards and reasoning logic among different individuals, making it challenging to establish a standardized and consistent reasoning process.

The emergence of cloud LLM services (denoted as $\mathcal{M}_{s}$) addresses the above issues. First, although the cloud LLM services provided by large model service providers may not be directly deployed in actual businesses, they can still generate some of reasoning data with a few API calls. Second, the judgment criteria of LLMs are unified, ensuring consistency in reasoning processes across different samples and maintaining the standardization of these processes. We can take advantage of cloud LLM services to generate reasoning processes for augmenting our training set. Subsequently, using the training set with reasons, we can fine-tune the open-source LLMs for specific domain tasks, thereby enhancing the capabilities of privately deployed models ($\mathcal{M}_{p}$) for practical real-world tasks.

\textbf{Weak supervision.} To obtain the reasoning process why a training sample $x_i$ is classified as class $y_i$, a natural approach is to prompt the $\mathcal{M}_{s}$ with sentence $x_i$ and its corresponding label $y$, and then require the $\mathcal{M}_{s}$ to analyse the reasoning process $\hat{r}_i$. However, previous studies have highlighted the issue of hallucinations in LLMs~\citep{manakul2023selfcheckgpt,bang2023multitask}. If $\mathcal{M}_{s}$ is given access to the true label before analyse, its response may be influenced by hallucinations. This would significantly compromise the quality of the data for fine-tuning the model. Consequently, we propose a weak supervision approach in which $\mathcal{M}_{s}$ is not provided with the true label $y_i$ during its reasoning process. Instead, it generates predicted labels $\hat{y}_i$ based on its reasoning process, and the true labels from the training set are only used to assess whether the judgments of $\mathcal{M}_{s}$ are correct.

Specifically, we can achieve this through the following steps:

\begin{enumerate}
\item Determine the instruction that includes the reasoning process based on the actual task and training dataset $\mathcal{D}_{raw}=\left\{x_i, y_i\right\}_{i=1}^N$.
\item Prompt the instruction and training data $x_i$ to a remote model service $\mathcal{M}_s$ to obtain the reasoning process $\hat{r_i}$ and predicted label $\hat{y_i}$.
\item Fine-tune the open-source model using the training set $\mathcal{D}_{train}=\left\{x_i, \hat{r}_i, \hat{y}_i\right\}_{i=1}^N$ that includes reasoning to get privately deployed model $\mathcal{M}_{p}$.
\end{enumerate}

However, although the model capabilities offered by LLM service providers surpass those of open-source models, these cloud model services might incorrectly classify the training data, which makes it difficult to acquire accurate reasons for training privately deployed models. 
To address this issue, we propose various strategies to clean the dataset when an incorrect response is obtained. Specifically, we select samples in which the $\mathcal{M}_s$ predicts inaccurately by examining the consistency between the predicted results and the ground truth.  
For samples with prediction errors, that is, samples where $\hat{y} \neq y$, we have two distinct strategies for managing them.

(a) Removing samples that are misclassified by $\mathcal{M}_{s}$, which is a simple and natural solution. The specific process involves inserting the following step between the step 2 and step 3:
\begin{enumerate}
\item[2.a.] If $\hat{y}_i\neq y_i$, remove the sample $<x_i, \hat{r}_i, \hat{y}_i>$ from the training dataset $\mathcal{D}_{train}$.
\end{enumerate}

(b) Introducing self-recheck to rectify the model's outputs. The specific process entails incorporating the subsequent step between steps 2 and 3:
\begin{enumerate}
\item[2.b.] If $\hat{y}_i\neq y_i$, prompt $\mathcal{M}_{s}$ to recheck the response with the original question and the incorrect reason $\hat{r}_i$ and predicted label $\hat{y}_i$, then update the revised reason $\hat{r}_i$ and the prediction $\hat{y}_i$.
\end{enumerate}
Though the self-recheck process can be repeated multiple times to obtain the correct prediction result, we perform the reflection once during the experiment. If $\mathcal{M}_{s}$ persists in making errors even after re-answering, it indicates the question is too hard for $\mathcal{M}_{s}$, and compelling it to change the response might result in lower-quality answers due to hallucinations. 

\textbf{Reason consistency checking} The consistency of the reasoning process is also crucial for fine-tuning the model, apart from weak supervision. For example, when reusing the CoT process to generate training data using $\mathcal{M}_s$, $\mathcal{M}_s$ may generate thought processes that do not match the final classification results (please refer to the appendix for a case study). By analyzing the consistency of the CoT process based on weak supervision, we can filter out contradictory samples in the review process, thereby improving the quality of the training samples. One specific approach is to break down the reasoning process into multiple steps and evaluate whether there are any contradictions between each step.

\subsection{Overfitting Analyze and Augmentation}

Although introducing CoT process can effectively prevent the model from overfitting during the fine-tuning stage, models fine-tuned on limited data still inevitably learn undesired decision shortcuts. These shortcuts can severely impact the model's performance and prevent it from adapting to open deployment environments. By incorporating the overfitting paradigm analysis strategy to capture the potential shortcuts learned by the model, and introducing data quality control strategy, we can effectively enhance the robustness of the model when deployed in open environments. Specifically, after fine-tuning the model on the training set, we need a validation set $\mathcal{D}_{val}=\{x'_i, y'_i\}_{i}^{N_\text{val}}$, which can be obtained from publicly available data without the need of including the analysis process $r'$. Then we evaluate the fine-tuned model on these samples, identifying the failure cases (i.e., $y'_i \neq \hat{y}'_i$). The LLM analyzes these failure cases to identify the shortcut paradigm causing them with the reasoning process $\hat{r}'$. Then, we use the LLM to generate adversarial samples that incorporate the identified shortcuts, and add them to the training set. The model is then fine-tuned again with the updated training set to eliminate and overcome these shortcuts. In this paper, we only conduct label consistency checking and approximate reason consistency checking by regenerating the reasoning processes (recheck) when there is an inconsistency in label. Recheck can help $\mathcal{M}_s$ replace the wrong reasoning process with a new one, which has the potential to replace it with a consistent reasoning process. More reason consistency checking will be deployed in the future work.

\subsection{Difference Between the Discriminative and Generative Content Moderation Model}

In this paper, we introduce a paradigm for a generative content moderation model to achieve decent performance using limited data. Compared to discriminative content moderation models, the generative models have several advantages: (1) a more flexible requirement for the quality of the training set, (2) reduced occurrence of undesired prediction shortcuts, (3) higher interpretability, and (4) no longer relying on high-quality manual annotations and adversarial methods.
$\bullet$ The strategy of weak supervision proposed in this paper allows for automated filtering of the dataset. Through the generative reasoning process, low-quality samples are filtered using label consistency checks and reasoning process consistency checks. This approach reduces the strict dependence on the dataset without requiring additional human recheck. 
$\bullet$ Generative models can effectively avoid overfitting on the training set, thereby reducing undesired decision shortcuts. In contrast to discriminative models, generative models do not simply match the input content to the final label class. Instead, they determine the risk types through a step-by-step deduction process. This deductive reasoning approach effectively avoids decision shortcuts, thereby enhancing the model's generalization ability. 
$\bullet$Additionally, generative models can provide a reasoning process as output, enhancing interpretability. Unlike discriminative models that only provide classification results and probabilities, generative models can output a detailed reasoning process in natural language format. This step-by-step analysis explains the reasons behind the final decision. It can pinpoint specific violations within the inappropriate content and provide richer background knowledge as feedback.
$\bullet$ Benefiting from the reasoning process, the generative model no longer relys on high-quality manual annotations and adversarial methods, making it easier to analyze the shortcut patterns of the model. For instance, the detailed deduction process obtained through fine-tuning the model can be statistically analyzed using human or more powerful language models (LLMs). This allows for a more accurate understanding of the model's shortcut patterns, enabling automated adversarial model modifications such as generating adversarial samples or model editing.

\section{Experiments}
\subsection{Experiments Setting}

\textbf{Dataset:} The business scenario in the experiment focuses on content moderation in a Chinese environment. We classify the content into six categories: Political Harmful (Political), Pornography (Porno), Violence, Discrimination or Insult (Offensive), Gambling, and Harmless. The data originates from our real-world business scenarios and public data that has been annotated with labels $y$ by human annotators. For example, samples of Offensive are taken from the open benchmark dataset COLD~\citep{deng2022cold}, while the Political Harmful items are collected from business data. The distribution of the data is shown in Table~\ref{tab:datasets}, where "total" represents the total number of data collected in each category. Initially, we removed duplicates from each category and retained the most diverse 1450 samples. Subsequently, we divided them into a training set of 1200 samples and a test set of 250 samples. To comprehensively evaluate the performance of the fine-tuned model, we also conducted validation on out-of-distribution (OOD) data. In the experiment, we selected HateSpeech~\cite{Cooke2022} as the OOD dataset (cross-lingual), which consists of three parts based on the platform source (Reddit, Twitter, 4Chan). Additionally, we tested the model's zero-shot capability on OOD tasks by deploying it to detect successful jailbreak attacks on large models (OOD harmful categories). We assessed the performance of the fine-tuned model using a zero-shot approach to determine if the large model produced harmful content, including categories that did not appear in the training set.

\textbf{Model:} For open-source base models, we have chosen Baichuan-7B-Chat and Baichuan-13B-Chat as the base models because they have demonstrated good performance in basic Chinese tasks and exhibit a more prominent understanding of the Chinese language context compared to LLaMA~\citep{touvron2023llama}. Although some researchers have aligned LLaMA for Chinese\footnote{\href{https://github.com/ymcui/Chinese-LLaMA-Alpaca}{https://github.com/ymcui/Chinese-LLaMA-Alpaca} \\ \href{https://huggingface.co/shibing624/chinese-llama-plus-13b-hf}{\, https://huggingface.co/shibing624/chinese-llama-plus-13b-hf}}, they lack official formal evaluations and guarantees, making them unsuitable as base models.

\textbf{Fine-tuning:} Fine-tuning the model can be done using methods including full-parameter-tuning, P-tuning~\citep{liu2022p}, or LORA~\citep{hu2021lora}. Parameter-Efficient Fine-Tuning (PEFT) has been proven effective in preventing overfitting in models, and several studies have demonstrated that PEFT outperforms full-parameter-tuning~\citep{zeng2023tim,peft}.Tuning the Baichuan-Chat based on LORA is among the widely recommended PEFT methods, so we fine-tune the Baichuan-Chat based on LORA.

\begin{table*}[ht]
\caption{Results on the different models fine-tuned with different strategies, where $\textcolor{red}{\blacktriangle}$ and $\textcolor{mycolor2}{\blacktriangle}$ indicate our fine-tuned models outperform GPT-3.5 and GPT-4 respectively, and ``Average'' indicates the micro-average performance on in-distribution dataset, the best performance in terms of F1 Score for different model sizes are \best{highlighted}.}\label{tab:allperformance}
\begin{center}
\resizebox{\textwidth}{!}{\begin{tabular}{ccc|ccccccc|ccc}
\toprule
\toprule
\multirow{2}{*}{\textbf{Model}}&\multirow{2}{*}{\textbf{Strategies}}&\multirow{2}{*}{\textbf{Metrics}} & \multicolumn{7}{c}{\textbf{In-distribution}} & \multicolumn{3}{c}{\textbf{OOD (Cross-lingual)}}\\
&&&  \improvedB{Political}& \improvedB{Porno} & \improvedB{Violent} & \improvedB{Offen.} & \improvedB{Gamb.} & \improvedB{Harml.} & \improvedB{Average} & \improvedA{Reddit} & \improvedA{Twitter} & \improvedA{4Chan}\\
\midrule

\multirow{3}{*}{\textbf{GPT-3.5}}&\multirow{3}{*}{-}&Recall&60.6&77.9&94.3&85.5&83.9&57.2&75.6&83.7&86.4&84.0 \\
&&Precision&40.7&76.2&54.3&46.8&76.4&40.3&52.8&78.4&79.2&81.6 \\
&&F1 Score&48.7&77.1&68.9&60.5&80.0&47.3&62.2&77.6&80.3&81.5 \\
\midrule
\multirow{3}{*}{\textbf{GPT-4}}&\multirow{3}{*}{-}&Recall&51.6&85.2&83.6&80.4&89.2&77.2&77.9&92.5&93.4&93.4 \\
&&Precision&67.9&92.6&76.3&53.6&96.5&45.0&67.6&88.6&90.4&93.0 \\
&&F1 Score&58.6&88.7&79.8&64.3&92.7&56.8&72.3 &89.9&91.6&93.0\\

\midrule
\midrule
\multirow{17}{*}{\begin{tabular}{c}
     \textbf{Baichuan2} \\
     \textbf{(7B)}
\end{tabular}}&\multirow{3}{*}{\textbf{Original}}

&Recall&61.2&48.8&42.0&57.6&56.0&34.0&49.9&46.9&54.1&46.5 \\
&&Precision&26.1&76.2&59.7&33.5&81.9&30.1&41.5&73.7&73.8&74.3 \\
&&F1 Score&36.6&59.5&49.3&42.4&66.5&32.0&45.3&52.9&55.6&52.0 \\

\cmidrule{2-13}

&\multirow{3}{*}{\textbf{Setting A}}&Recall&51.6&83.6&83.2&74.8&87.2&74.4&75.8&81.6&86.7&90.3 \\
&&Precision&69.4&93.3&76.2&55.8&96.0&42.8&67.7&83.7&84.9&89.4 \\
&&F1 Score&59.2&88.2&79.5&63.9&91.4&54.3&71.5 &82.0&84.7&\best{89.4}\\

\cmidrule{2-13}

&\multirow{3}{*}{\textbf{Setting B}}&Recall&66.8&85.6&94.0&81.2&90.4&72.8&81.8&83.0&85.9&88.5 \\
&&Precision&71.4&90.3&78.9&55.5&97.4&52.3&71.5&86.3&85.2&87.2 \\
&&F1 Score&69.0&87.9&\best{85.8}&65.9&\best{93.8}&60.9&76.3&\best{84.3}&84.2&87.3 \\

\cmidrule{2-13}

&\multirow{3}{*}{\textbf{Setting C}}&Recall&62.0&83.2&90.0&75.6&87.2&78.4&79.4&81.1&86.3&88.5 \\
&&Precision&72.8&92.0&77.1&58.0&97.3&47.7&70.4&85.7&85.7&88.6 \\
&&F1 Score&67.0&87.4&83.0&65.6&92.0&59.3&74.6&83.0&\best{84.8}&88.2 \\

\cmidrule{2-13}

&\multirow{3}{*}{\textbf{Setting D}}&Recall&68.4&85.6&92.8&79.2&91.2&74.0&81.9&79.9&83.4&87.5 \\
&&Precision&73.1&94.3&77.3&56.9&96.6&52.6&72.4&83.5&84.2&86.5 \\
&&F1 Score&\best{70.7}&\best{89.7}&{84.4}&\best{66.2}&\best{93.8}&\best{61.5}&\best{76.8}&81.0&82.6&86.3 \\

\midrule
\midrule
\multirow{17}{*}{\begin{tabular}{c}
     \textbf{Baichuan2} \\
     \textbf{(13B)}
\end{tabular}}&\multirow{3}{*}{\textbf{Original}}&Recall&74.4&76.4&58.0&90.4&78.8&30.8&68.1&66.2&63.5&65.4 \\
&&Precision&23.8&68.7&41.9&31.4&73.2&32.4&38.8 &68.1&66.5&69.6\\
&&F1 Score&36.1&72.3&48.7&46.6&75.9&31.6&49.5 &56.8&51.4&57.2\\

\cmidrule{2-13}

&\multirow{3}{*}{\textbf{Setting A}}&Recall&51.2&83.2&83.2&76.0&87.2&76.4&76.2&83.0&85.7&88.1 \\
&&Precision&68.4&92.9&77.9&56.0&98.2&42.7&67.8&85.0&85.9&89.1 \\
&&F1 Score&58.6&87.8&80.5&64.5&92.4&54.8&71.8&83.7&84.8&88.4 \\

\cmidrule{2-13}
&\multirow{3}{*}{\textbf{Setting B}}&Recall&68.4&86.4&90.4&80.8&93.2&75.2&82.4&82.4&85.4&87.6 \\
&&Precision&70.4&90.4&76.1&55.2&96.3&52.4&70.8&85.2&83.5&88.1 \\
&&F1 Score&69.4&88.3&82.6&\best{65.6}&94.7&61.7&76.2&83.3&83.2&87.5\\

\cmidrule{2-13}

&\multirow{3}{*}{\textbf{Setting C}}&Recall&66.4&85.6&88.0&74.4&89.6&75.6&79.9&84.6&86.7&89.3 \\
&&Precision&76.5&91.1&78.0&53.4&97.4&47.6&70.2&86.1&85.5&89.4 \\
&&F1 Score&71.1&88.2&82.7&62.2&93.3&58.4&74.7&\best{85.0}&\best{84.9}&\best{89.1}\\

\cmidrule{2-13}

&\multirow{3}{*}{\textbf{Setting D}}&Recall&72.8&86.8&93.2&76.8&94.0&76.0&83.3&83.2&84.4&88.2 \\
&&Precision&70.5&91.2&77.2&52.6&95.9&55.6&71.4&83.7&83.3&86.7 \\
&&F1 Score&\best{71.7}&\best{88.9}&\best{84.4}&{62.4}&\best{94.9}&\best{64.2}&\best{76.9}&{82.7}&{82.4}&{86.9}\\

\bottomrule
\bottomrule
\end{tabular}}
\end{center}
\end{table*}
\begin{table*}[ht]
\caption{Results on the different models without reasoning processes during deployment.}\label{tab:trainwithcotinferencewocot}
\begin{center}
\resizebox{\textwidth}{!}{\begin{tabular}{ccc|ccccccc|ccc}
\toprule

\multirow{2}{*}{\textbf{Model}}&\multirow{2}{*}{\textbf{Strategies}}&\multirow{2}{*}{\textbf{Metrics}} & \multicolumn{7}{c}{\textbf{In-distribution}} & \multicolumn{3}{c}{\textbf{OOD (Cross-lingual)}}\\
&&&  {Political}& {Porno} &{Violent} & {Offen.} & {Gamb.} & {Harml.} & {Average} & {Reddit} & {Twitter} & {4Chan}\\
\midrule

\multirow{6}{*}{\begin{tabular}{c}
     \textbf{Baichuan2} \\
     \textbf{(7B)}
\end{tabular}}&\multirow{3}{*}{\textbf{Original}}

&Recall&66.8&19.6&5.2&25.6&0.8&40.0&26.3 &33.0&28.2&30.5\\
&&Precision&21.2&90.7&76.5&65.3&50.0&21.5&27.7&81.9&80.9&81.3 \\
&&F1 Score&32.2&32.2&9.7&36.8&1.6&28.0&27.0&46.9&41.7&44.3 \\

\cmidrule{2-13}

&\multirow{3}{*}{\textbf{Setting B}}&Recall&80.4&92.0&98.4&80.4&88.4&67.2&84.5&84.3&85.3&89.5 \\
&&Precision&59.3&78.0&75.9&51.9&88.8&54.0&66.5&87.7&82.3&88.9 \\
&&F1 Score&68.3&84.4&85.7&63.1&88.6&59.9&74.4 &85.6&82.0&88.9\\

\midrule
\multirow{6}{*}{\begin{tabular}{c}
     \textbf{Baichuan2} \\
     \textbf{(13B)}
\end{tabular}}&\multirow{3}{*}{\textbf{Original}}&Recall&61.6&77.2&67.6&98.0&68.0&28.0&66.7&71.5&66.8&76.2 \\
&&Precision&48.7&91.5&83.7&44.2&93.4&43.8&61.6 &74.7&70.4&78.3 \\
&&F1 Score&54.4&83.7&74.8&60.9&78.7&34.1&64.1 &65.2&55.0&71.3 \\

\cmidrule{2-13}
&\multirow{3}{*}{\textbf{Setting B}}&Recall&83.2&92.0&97.2&86.4&92.4&66.0&86.2&81.5&82.9&87.4 \\
&&Precision&56.1&77.2&73.6&63.0&91.3&61.1&69.3&90.2&86.2&92.0 \\
&&F1 Score&67.0&83.9&83.8&72.8&91.8&63.5&76.8&85.3&83.3&89.5 \\

\bottomrule

\end{tabular}}
\end{center}
\end{table*}
\textbf{Different training strategies.}
We apply various fine-tuning strategies, such as reject strategy (detials are shown in Setting B), recheck strategy, and a combination of both strategies (for the prompt templates please refer to Appendix~\ref{sec:app-instruction}).

\textbf{GPT-3.5 \& GPT-4:} Ideally, we hope the privately deployed model to achieve competitive performance comparable to the cloud LLM service. To evaluate this, we test the zero-shot performance of GPT-3.5-turbo and GPT-4 using the same instructions, which involves conducting content analysis before making a judgment. \emph{\textbf{Please note that} OpenAI has its content management policy. Some content is filtered by content filtering policies, and in such cases, we consider it as a successful detection by GPT if the responses of positive samples are filtered.}  

\textbf{Original:} We evaluate the Baichuan-7B-Chat and Baichuan-13B-Chat models without any fine-tuning for content moderation as the baseline, to offer an intuitive comparison of the fine-tuning effects achieved through various strategies.

\textbf{Setting A:} We generate CoT $\hat{r}$ and predicted class $\hat{y}$ using GPT-4 on the training set. Then, we utilize $\mathcal{D}_{train}=\left\{x_i, \hat{r}_i, \hat{y}_i\right\}_{i=1}^N$ to fine-tune Baichuan-7B-Chat and Baichuan-13B-Chat.

\textbf{Setting B:} Taking into account that the answers generated by GPT-4 may be incorrect, we check GPT-4's responses by the ground truth. We enhance the quality of the training data using discarding samples for which GPT-4 made inaccurate predictions (i.e., $\hat{y}\neq y$), as demonstrated in step 2.a.

\textbf{Setting C:} In addition to discarding samples with incorrect predictions, we instruct GPT-4 to perform self-rechecks and modify its responses for these samples as shown in step 2.b.

\textbf{Setting D:} This strategy combines the recheck and reject approaches. Firstly, the model reflects on and modifies incorrect answers. Subsequently, samples that persist as incorrect after re-answering are removed from the training dataset.

\subsection{Results}\label{sec:results}

We conducted a comprehensive evaluation of all models, including the original model. The results are presented in Table~\ref{tab:allperformance}. Notably, we highlight the models that outperform GPT-3.5 or GPT-4 in each domain. Upon analyzing Table~\ref{tab:allperformance}, it becomes evident that while these models initially perform poorly before fine-tuning, all fine-tuned models exhibit improved performance compared to GPT-3.5. Particularly, the performance of GPT-4 significantly surpasses that of GPT-3.5, especially in the Harmless domain.
Interestingly, even before fine-tuning, the 13B model slightly outperforms the 7B model, and this trend persists after fine-tuning, with the 13B model demonstrating slight improvements over the 7B model. Furthermore, incorporating weak-supervision and recheck strategies leads to a substantial performance improvement of approximately 5\%, regardless of the base model's size. Notably, when both measures are applied simultaneously (Setting D), the model's performance reaches its optimum.
In specific domains, the most significant improvement is observed in the Harmless domain, where both the 7B and 13B models outperform GPT-4 after implementing the weak-supervision and recheck strategies. Additionally, we assessed the cross-lingual out-of-domain (OOD) generalization capability of the fine-tuned models. We conducted tests on Offensive data across three different open-source platforms (Reddit, Twitter, 4Chan), and the results indicate that our model surpasses GPT-3.5 and performs significantly better than the non-fine-tuned model. This demonstrates that the model acquires a certain level of cross-lingual content moderation ability after undergoing fine-tuning.

\begin{figure}
  \centering
  \includegraphics[width=0.48\textwidth]{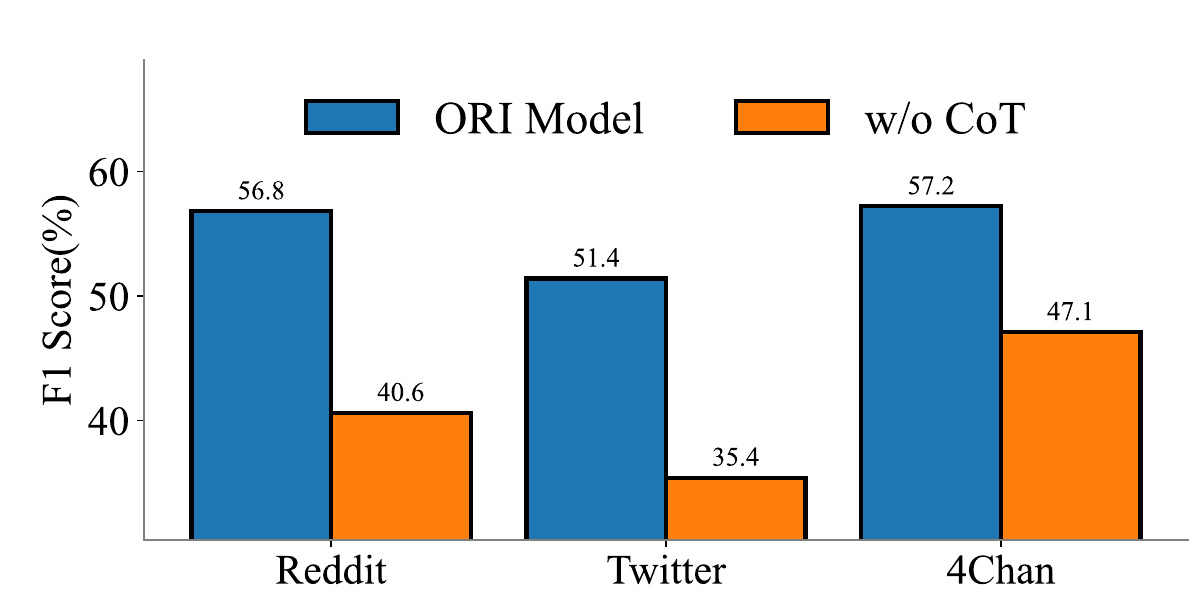}
 \caption{F1 Score on OOD datasets.}
 \label{fig:ood_wocot}
\end{figure}
\textbf{Fine-tuning LLMs with the pattern $(x_i, y_i)$ can lead to severe overfitting.} As shown in Fig.~\ref{fig:ood_wocot}, after fine-tuning without incorporating reasons, the model's performance significantly deteriorates on out-of-distribution (OOD) data. Specifically, Fig.~\ref{fig:ood_wocot} displays the recall rates on negative samples of different models on binary classification OOD benchmark data. In the setting without fine-tuning the reasoning process, the 13B model, after being fine-tuned, exhibits a noticeable prediction bias, where it predicts all samples in the OOD data as belonging to the same category. In other words, the model fine-tuned without incorporating reasons exhibits severe overfitting (details please refer to Appendix~\ref{sec:app-wocot}), making it unsuitable for deployment in practical business scenarios.

\textbf{Weak supervision can improve the fine-tuning performance.}
Compared to the unsupervised setting (Setting A), we introduce weak supervision methods to improve the quality of reasoning processes in the training set. Specifically, we find that (1) filtering low-quality reasoning processes through weak supervision can help enhance fine-tuning performance. As shown in Table~\ref{tab:allperformance} (Setting B), the model's performance significantly improves after filtering out low-quality samples. (2) Rechecking low-quality reasoning processes through weak supervision can also enhance the quality of the fine-tuning data. As shown in Table~\ref{tab:allperformance} (Setting C), the model's performance improves by regenerating reasoning processes for low-quality samples. Moreover, these two strategies are not conflicting. When both strategies are combined (Setting D), the model achieves similar or even better results than the best performance of either strategy alone (for the reasoning processes generated from GPT-4 please refer to Appendix~\ref{sec:app_case}).

\textbf{Fine-tuning with reasoning processes improve models inherently.} In addition to the results in Table~\ref{tab:allperformance}, we also found that the model exhibits a significant improvement even when deployed without outputting reasoning processes, after fine-tuning the model with accompanying reasoning processes. Furthermore, as shown in Table~\ref{tab:trainwithcotinferencewocot}, the results generated by the model when deployed without outputting reasoning processes are almost indistinguishable from the results with outputted reasoning processes. This indicates that the model inherently overcomes decision shortcuts and overfitting paradigms. This is crucial for fine-tuning vertical domain models because outputting reasoning processes during deployment implies higher inference costs and slower response times. However, if the model only needs to introduce reasoning processes during fine-tuning and does not require reasoning processes during deployment to generate accurate answers, the cost introduced by reasoning processes is only applied during the training phase. This allows for more flexible deployment of the model while ensuring testing performance.

\begin{table}[ht]
\caption{Results on OOD task with zero-shot.}\label{tab:zero-shot}
\begin{center}
{\begin{tabular}{c|ccc}
\toprule

\multirow{2}{*}{\textbf{Categories}}&\multirow{2}{*}{\textbf{GPT-4}}&\multicolumn{2}{c}{\textbf{Ours}}\\
 &  & w/o CoT & with CoT \\
\midrule

Political & 70.0 & \best{86.6} & \best{86.6} \\
Porno & \best{87.0} &73.3 & 86.0 \\
Offen. & 83.0 &\best{86.0} &73.3 \\
Violent & \best{93.0} &86.0& 86.0 \\
\midrule
Internet Fraud & 83.0 &\best{86.6}& \best{86.6} \\
Privacy Disclosure & \best{88.0}&66.0 & 80.0 \\
Illegal Suggestion & \best{93.0} &86.0 & \best{93.0} \\

\bottomrule

\end{tabular}}
\end{center}
\end{table}
\textbf{The fine-tuned model shows zero-shot capability on new tasks.}  In the results shown in Table~\ref{tab:allperformance}, we evaluated the model's cross-lingual capability. In addition, we also assessed the model's zero-shot capability on out-of-distribution tasks in the context of Chinese content moderation. Specifically, we deployed the model for detecting jailbreak attacks on large models. This means that the fine-tuned model was used to determine if the model's output was successfully manipulated to produce inappropriate content, with the large model's responses serving as test samples. In Table~\ref{tab:zero-shot}, we present the zero-shot classification results for categories such as Internet Fraud, Privacy Disclosure, and Illegal Suggestion (categories that did not appear in the training set). From the results, it can be observed that the model performs competitively on these tasks, comparable to GPT-4. This indicates that the model can generalize to other tasks beyond the training set after fine-tuning.



\section{Conclusion}
In this paper, we conduct a fine-tuning attempt to privately deploy LLM with content moderation as a specific scenario. We enhance the model's generalization ability by introducing reasoning process during the fine-tuning process, resulting in the privately deployed model achieving performance that is better than that of GPT-3.5 on content moderation. Additionally, we introduce weak supervision to improve the reasoning quality of the training dataset, which improve the fine-tuning performance at a small annotation cost. We also discuss the different design choice of data engineering and SFT. Overall, we believe the research offers valuable insights for researchers in various fields to fine-tune privately deployed models in their domains.

\bibliographystyle{ACM-Reference-Format}
\bibliography{sample-base}


\begin{thebibliography}{47}


\ifx \showCODEN    \undefined \def \showCODEN     #1{\unskip}     \fi
\ifx \showDOI      \undefined \def \showDOI       #1{#1}\fi
\ifx \showISBNx    \undefined \def \showISBNx     #1{\unskip}     \fi
\ifx \showISBNxiii \undefined \def \showISBNxiii  #1{\unskip}     \fi
\ifx \showISSN     \undefined \def \showISSN      #1{\unskip}     \fi
\ifx \showLCCN     \undefined \def \showLCCN      #1{\unskip}     \fi
\ifx \shownote     \undefined \def \shownote      #1{#1}          \fi
\ifx \showarticletitle \undefined \def \showarticletitle #1{#1}   \fi
\ifx \showURL      \undefined \def \showURL       {\relax}        \fi
\providecommand\bibfield[2]{#2}
\providecommand\bibinfo[2]{#2}
\providecommand\natexlab[1]{#1}
\providecommand\showeprint[2][]{arXiv:#2}

\bibitem[Bang et~al\mbox{.}(2023)]%
        {bang2023multitask}
\bibfield{author}{\bibinfo{person}{Yejin Bang}, \bibinfo{person}{Samuel Cahyawijaya}, \bibinfo{person}{Nayeon Lee}, \bibinfo{person}{Wenliang Dai}, \bibinfo{person}{Dan Su}, \bibinfo{person}{Bryan Wilie}, \bibinfo{person}{Holy Lovenia}, \bibinfo{person}{Ziwei Ji}, \bibinfo{person}{Tiezheng Yu}, \bibinfo{person}{Willy Chung}, {et~al\mbox{.}}} \bibinfo{year}{2023}\natexlab{}.
\newblock \showarticletitle{A multitask, multilingual, multimodal evaluation of chatgpt on reasoning, hallucination, and interactivity}.
\newblock \bibinfo{journal}{\emph{arXiv preprint arXiv:2302.04023}} (\bibinfo{year}{2023}).
\newblock


\bibitem[Bhardwaj and Poria(2023)]%
        {bhardwaj2023red}
\bibfield{author}{\bibinfo{person}{Rishabh Bhardwaj} {and} \bibinfo{person}{Soujanya Poria}.} \bibinfo{year}{2023}\natexlab{}.
\newblock \showarticletitle{Red-Teaming Large Language Models using Chain of Utterances for Safety-Alignment}.
\newblock \bibinfo{journal}{\emph{arXiv preprint arXiv:2308.09662}} (\bibinfo{year}{2023}).
\newblock


\bibitem[Cooke(2022)]%
        {Cooke2022}
\bibfield{author}{\bibinfo{person}{Shane Cooke}.} \bibinfo{year}{2022}\natexlab{}.
\newblock \showarticletitle{{Labelled Hate Speech Detection Dataset.}}
\newblock  (\bibinfo{date}{4} \bibinfo{year}{2022}).
\newblock
\urldef\tempurl%
\url{https://doi.org/10.6084/m9.figshare.19686954.v1}
\showDOI{\tempurl}


\bibitem[Dan et~al\mbox{.}(2023)]%
        {educhat2023}
\bibfield{author}{\bibinfo{person}{Yuhao Dan}, \bibinfo{person}{Zhikai Lei}, \bibinfo{person}{Gu Yiyang}, \bibinfo{person}{Yong Li}, \bibinfo{person}{Jianghao Yin}, \bibinfo{person}{Jiaju Lin}, \bibinfo{person}{Linhao Ye}, \bibinfo{person}{Zhiyan Tie}, \bibinfo{person}{Yougen Zhou}, \bibinfo{person}{Yilei Wang}, \bibinfo{person}{Aimin Zhou}, \bibinfo{person}{Ze Zhou}, \bibinfo{person}{Qin Chen}, \bibinfo{person}{Jie Zhou}, \bibinfo{person}{Liang He}, {and} \bibinfo{person}{Xipeng Qiu}.} \bibinfo{year}{2023}\natexlab{}.
\newblock \showarticletitle{EduChat: A Large-Scale Language Model-based Chatbot System for Intelligent Education}.
\newblock \bibinfo{journal}{\emph{arXiv preprint arXiv:2308.02773}} (\bibinfo{year}{2023}).
\newblock


\bibitem[Deng et~al\mbox{.}(2022)]%
        {deng2022cold}
\bibfield{author}{\bibinfo{person}{Jiawen Deng}, \bibinfo{person}{Jingyan Zhou}, \bibinfo{person}{Hao Sun}, \bibinfo{person}{Chujie Zheng}, \bibinfo{person}{Fei Mi}, \bibinfo{person}{Helen Meng}, {and} \bibinfo{person}{Minlie Huang}.} \bibinfo{year}{2022}\natexlab{}.
\newblock \showarticletitle{Cold: A benchmark for chinese offensive language detection}.
\newblock \bibinfo{journal}{\emph{arXiv preprint arXiv:2201.06025}} (\bibinfo{year}{2022}).
\newblock


\bibitem[Dinan et~al\mbox{.}(2019)]%
        {dinan2019build}
\bibfield{author}{\bibinfo{person}{Emily Dinan}, \bibinfo{person}{Samuel Humeau}, \bibinfo{person}{Bharath Chintagunta}, {and} \bibinfo{person}{Jason Weston}.} \bibinfo{year}{2019}\natexlab{}.
\newblock \showarticletitle{Build it break it fix it for dialogue safety: Robustness from adversarial human attack}.
\newblock \bibinfo{journal}{\emph{arXiv preprint arXiv:1908.06083}} (\bibinfo{year}{2019}).
\newblock


\bibitem[Du et~al\mbox{.}(2022)]%
        {du2022glm}
\bibfield{author}{\bibinfo{person}{Zhengxiao Du}, \bibinfo{person}{Yujie Qian}, \bibinfo{person}{Xiao Liu}, \bibinfo{person}{Ming Ding}, \bibinfo{person}{Jiezhong Qiu}, \bibinfo{person}{Zhilin Yang}, {and} \bibinfo{person}{Jie Tang}.} \bibinfo{year}{2022}\natexlab{}.
\newblock \showarticletitle{GLM: General Language Model Pretraining with Autoregressive Blank Infilling}. In \bibinfo{booktitle}{\emph{Proceedings of the 60th Annual Meeting of the Association for Computational Linguistics (Volume 1: Long Papers)}}. \bibinfo{pages}{320--335}.
\newblock


\bibitem[Franco et~al\mbox{.}(2023)]%
        {franco2023analyzing}
\bibfield{author}{\bibinfo{person}{Mirko Franco}, \bibinfo{person}{Ombretta Gaggi}, {and} \bibinfo{person}{Claudio~E Palazzi}.} \bibinfo{year}{2023}\natexlab{}.
\newblock \showarticletitle{Analyzing the Use of Large Language Models for Content Moderation with ChatGPT Examples}. In \bibinfo{booktitle}{\emph{3rd International Workshop on Open Challenges in Online Social Networks}}. \bibinfo{pages}{1--8}.
\newblock


\bibitem[Gao et~al\mbox{.}(2023)]%
        {gao2023pal}
\bibfield{author}{\bibinfo{person}{Luyu Gao}, \bibinfo{person}{Aman Madaan}, \bibinfo{person}{Shuyan Zhou}, \bibinfo{person}{Uri Alon}, \bibinfo{person}{Pengfei Liu}, \bibinfo{person}{Yiming Yang}, \bibinfo{person}{Jamie Callan}, {and} \bibinfo{person}{Graham Neubig}.} \bibinfo{year}{2023}\natexlab{}.
\newblock \showarticletitle{Pal: Program-aided language models}. In \bibinfo{booktitle}{\emph{International Conference on Machine Learning}}. PMLR, \bibinfo{pages}{10764--10799}.
\newblock


\bibitem[Gerrard(2018)]%
        {gerrard2018beyond}
\bibfield{author}{\bibinfo{person}{Ysabel Gerrard}.} \bibinfo{year}{2018}\natexlab{}.
\newblock \showarticletitle{Beyond the hashtag: Circumventing content moderation on social media}.
\newblock \bibinfo{journal}{\emph{New Media \& Society}} \bibinfo{volume}{20}, \bibinfo{number}{12} (\bibinfo{year}{2018}), \bibinfo{pages}{4492--4511}.
\newblock


\bibitem[Hajibabaee et~al\mbox{.}(2022)]%
        {hajibabaee2022offensive}
\bibfield{author}{\bibinfo{person}{Parisa Hajibabaee}, \bibinfo{person}{Masoud Malekzadeh}, \bibinfo{person}{Mohsen Ahmadi}, \bibinfo{person}{Maryam Heidari}, \bibinfo{person}{Armin Esmaeilzadeh}, \bibinfo{person}{Reyhaneh Abdolazimi}, {and} \bibinfo{person}{H James~Jr}.} \bibinfo{year}{2022}\natexlab{}.
\newblock \showarticletitle{Offensive language detection on social media based on text classification}. In \bibinfo{booktitle}{\emph{2022 IEEE 12th Annual Computing and Communication Workshop and Conference (CCWC)}}. IEEE, \bibinfo{pages}{0092--0098}.
\newblock


\bibitem[He et~al\mbox{.}(2010)]%
        {he2010data}
\bibfield{author}{\bibinfo{person}{Qinlu He}, \bibinfo{person}{Zhanhuai Li}, {and} \bibinfo{person}{Xiao Zhang}.} \bibinfo{year}{2010}\natexlab{}.
\newblock \showarticletitle{Data deduplication techniques}. In \bibinfo{booktitle}{\emph{2010 international conference on future information technology and management engineering}}, Vol.~\bibinfo{volume}{1}. IEEE, \bibinfo{pages}{430--433}.
\newblock


\bibitem[Hsieh et~al\mbox{.}(2023)]%
        {hsieh2023distilling}
\bibfield{author}{\bibinfo{person}{Cheng-Yu Hsieh}, \bibinfo{person}{Chun-Liang Li}, \bibinfo{person}{Chih-Kuan Yeh}, \bibinfo{person}{Hootan Nakhost}, \bibinfo{person}{Yasuhisa Fujii}, \bibinfo{person}{Alexander Ratner}, \bibinfo{person}{Ranjay Krishna}, \bibinfo{person}{Chen-Yu Lee}, {and} \bibinfo{person}{Tomas Pfister}.} \bibinfo{year}{2023}\natexlab{}.
\newblock \showarticletitle{Distilling step-by-step! outperforming larger language models with less training data and smaller model sizes}.
\newblock \bibinfo{journal}{\emph{arXiv preprint arXiv:2305.02301}} (\bibinfo{year}{2023}).
\newblock


\bibitem[Hu et~al\mbox{.}(2021)]%
        {hu2021lora}
\bibfield{author}{\bibinfo{person}{Edward~J Hu}, \bibinfo{person}{Yelong Shen}, \bibinfo{person}{Phillip Wallis}, \bibinfo{person}{Zeyuan Allen-Zhu}, \bibinfo{person}{Yuanzhi Li}, \bibinfo{person}{Shean Wang}, \bibinfo{person}{Lu Wang}, {and} \bibinfo{person}{Weizhu Chen}.} \bibinfo{year}{2021}\natexlab{}.
\newblock \showarticletitle{Lora: Low-rank adaptation of large language models}.
\newblock \bibinfo{journal}{\emph{arXiv preprint arXiv:2106.09685}} (\bibinfo{year}{2021}).
\newblock


\bibitem[Huang et~al\mbox{.}(2023)]%
        {lawyer-llama-report}
\bibfield{author}{\bibinfo{person}{Quzhe Huang}, \bibinfo{person}{Mingxu Tao}, \bibinfo{person}{Zhenwei An}, \bibinfo{person}{Chen Zhang}, \bibinfo{person}{Cong Jiang}, \bibinfo{person}{Zhibin Chen}, \bibinfo{person}{Zirui Wu}, {and} \bibinfo{person}{Yansong Feng}.} \bibinfo{year}{2023}\natexlab{}.
\newblock \showarticletitle{Lawyer LLaMA Technical Report}.
\newblock \bibinfo{journal}{\emph{ArXiv}}  \bibinfo{volume}{abs/2305.15062} (\bibinfo{year}{2023}).
\newblock


\bibitem[Husain and Uzuner(2021)]%
        {husain2021survey}
\bibfield{author}{\bibinfo{person}{Fatemah Husain} {and} \bibinfo{person}{Ozlem Uzuner}.} \bibinfo{year}{2021}\natexlab{}.
\newblock \showarticletitle{A survey of offensive language detection for the Arabic language}.
\newblock \bibinfo{journal}{\emph{ACM Transactions on Asian and Low-Resource Language Information Processing (TALLIP)}} \bibinfo{volume}{20}, \bibinfo{number}{1} (\bibinfo{year}{2021}), \bibinfo{pages}{1--44}.
\newblock


\bibitem[Jahan and Oussalah(2023)]%
        {jahan2023systematic}
\bibfield{author}{\bibinfo{person}{Md~Saroar Jahan} {and} \bibinfo{person}{Mourad Oussalah}.} \bibinfo{year}{2023}\natexlab{}.
\newblock \showarticletitle{A systematic review of Hate Speech automatic detection using Natural Language Processing.}
\newblock \bibinfo{journal}{\emph{Neurocomputing}} (\bibinfo{year}{2023}), \bibinfo{pages}{126232}.
\newblock


\bibitem[James et~al\mbox{.}(2023)]%
        {aws}
\bibfield{author}{\bibinfo{person}{Poquiz James}, \bibinfo{person}{Man Han}, {and} \bibinfo{person}{Tinaztepe Safa}.} \bibinfo{year}{2023}\natexlab{}.
\newblock \bibinfo{title}{AWS performs fine-tuning on a Large Language Model (LLM) to classify toxic speech for a large gaming company}.
\newblock \bibinfo{howpublished}{\url{https://aws.amazon.com/cn/blogs/machine-learning/}}.
\newblock


\bibitem[Kebriaei et~al\mbox{.}(2023)]%
        {kebriaei2023persian}
\bibfield{author}{\bibinfo{person}{Emad Kebriaei}, \bibinfo{person}{Ali Homayouni}, \bibinfo{person}{Roghayeh Faraji}, \bibinfo{person}{Armita Razavi}, \bibinfo{person}{Azadeh Shakery}, \bibinfo{person}{Heshaam Faili}, {and} \bibinfo{person}{Yadollah Yaghoobzadeh}.} \bibinfo{year}{2023}\natexlab{}.
\newblock \showarticletitle{Persian offensive language detection}.
\newblock \bibinfo{journal}{\emph{Machine Learning}} (\bibinfo{year}{2023}), \bibinfo{pages}{1--21}.
\newblock


\bibitem[Li et~al\mbox{.}(2023a)]%
        {li2023hot}
\bibfield{author}{\bibinfo{person}{Lingyao Li}, \bibinfo{person}{Lizhou Fan}, \bibinfo{person}{Shubham Atreja}, {and} \bibinfo{person}{Libby Hemphill}.} \bibinfo{year}{2023}\natexlab{a}.
\newblock \showarticletitle{``HOT" ChatGPT: The promise of ChatGPT in detecting and discriminating hateful, offensive, and toxic comments on social media}.
\newblock \bibinfo{journal}{\emph{arXiv preprint arXiv:2304.10619}} (\bibinfo{year}{2023}).
\newblock


\bibitem[Li et~al\mbox{.}(2023b)]%
        {chatglm-for-variable-star}
\bibfield{author}{\bibinfo{person}{YuYang Li}, \bibinfo{person}{CunShi Wang}, \bibinfo{person}{MengWei Qu}, \bibinfo{person}{Yu Bai}, \bibinfo{person}{Roberto Soria}, {and} \bibinfo{person}{JiFeng Liu}.} \bibinfo{year}{2023}\natexlab{b}.
\newblock \bibinfo{title}{StarGLM}.
\newblock \bibinfo{howpublished}{\url{https://github.com/Yu-Yang-Li/StarGLM}}.
\newblock


\bibitem[Liang et~al\mbox{.}(2023)]%
        {liang2023less}
\bibfield{author}{\bibinfo{person}{Chen Liang}, \bibinfo{person}{Simiao Zuo}, \bibinfo{person}{Qingru Zhang}, \bibinfo{person}{Pengcheng He}, \bibinfo{person}{Weizhu Chen}, {and} \bibinfo{person}{Tuo Zhao}.} \bibinfo{year}{2023}\natexlab{}.
\newblock \showarticletitle{Less is more: Task-aware layer-wise distillation for language model compression}. In \bibinfo{booktitle}{\emph{International Conference on Machine Learning}}. PMLR, \bibinfo{pages}{20852--20867}.
\newblock


\bibitem[Lightman et~al\mbox{.}(2023)]%
        {lightman2023lets}
\bibfield{author}{\bibinfo{person}{Hunter Lightman}, \bibinfo{person}{Vineet Kosaraju}, \bibinfo{person}{Yura Burda}, \bibinfo{person}{Harri Edwards}, \bibinfo{person}{Bowen Baker}, \bibinfo{person}{Teddy Lee}, \bibinfo{person}{Jan Leike}, \bibinfo{person}{John Schulman}, {and} \bibinfo{person}{Ilya Sutskever}.} \bibinfo{year}{2023}\natexlab{}.
\newblock \showarticletitle{Let’s Verify Step by Step}.
\newblock  (\bibinfo{year}{2023}).
\newblock


\bibitem[Lilian~Weng(2023)]%
        {lilian2023moderation}
\bibfield{author}{\bibinfo{person}{Andrea~Vallone Lilian~Weng, Vik~Goel}.} \bibinfo{year}{2023}\natexlab{}.
\newblock \showarticletitle{Using GPT-4 for content moderation}.
\newblock  (\bibinfo{year}{2023}).
\newblock


\bibitem[Liu et~al\mbox{.}(2021)]%
        {liu2021makes}
\bibfield{author}{\bibinfo{person}{Jiachang Liu}, \bibinfo{person}{Dinghan Shen}, \bibinfo{person}{Yizhe Zhang}, \bibinfo{person}{Bill Dolan}, \bibinfo{person}{Lawrence Carin}, {and} \bibinfo{person}{Weizhu Chen}.} \bibinfo{year}{2021}\natexlab{}.
\newblock \showarticletitle{What Makes Good In-Context Examples for GPT-$3 $?}
\newblock \bibinfo{journal}{\emph{arXiv preprint arXiv:2101.06804}} (\bibinfo{year}{2021}).
\newblock


\bibitem[Liu et~al\mbox{.}(2022)]%
        {liu2022p}
\bibfield{author}{\bibinfo{person}{Xiao Liu}, \bibinfo{person}{Kaixuan Ji}, \bibinfo{person}{Yicheng Fu}, \bibinfo{person}{Weng Tam}, \bibinfo{person}{Zhengxiao Du}, \bibinfo{person}{Zhilin Yang}, {and} \bibinfo{person}{Jie Tang}.} \bibinfo{year}{2022}\natexlab{}.
\newblock \showarticletitle{P-tuning: Prompt tuning can be comparable to fine-tuning across scales and tasks}. In \bibinfo{booktitle}{\emph{Proceedings of the 60th Annual Meeting of the Association for Computational Linguistics (Volume 2: Short Papers)}}. \bibinfo{pages}{61--68}.
\newblock


\bibitem[Luo et~al\mbox{.}(2023a)]%
        {luo2023wizardmath}
\bibfield{author}{\bibinfo{person}{Haipeng Luo}, \bibinfo{person}{Qingfeng Sun}, \bibinfo{person}{Can Xu}, \bibinfo{person}{Pu Zhao}, \bibinfo{person}{Jianguang Lou}, \bibinfo{person}{Chongyang Tao}, \bibinfo{person}{Xiubo Geng}, \bibinfo{person}{Qingwei Lin}, \bibinfo{person}{Shifeng Chen}, {and} \bibinfo{person}{Dongmei Zhang}.} \bibinfo{year}{2023}\natexlab{a}.
\newblock \showarticletitle{WizardMath: Empowering Mathematical Reasoning for Large Language Models via Reinforced Evol-Instruct}.
\newblock \bibinfo{journal}{\emph{arXiv preprint arXiv:2308.09583}} (\bibinfo{year}{2023}).
\newblock


\bibitem[Luo et~al\mbox{.}(2023b)]%
        {luo2023wizardcoder}
\bibfield{author}{\bibinfo{person}{Ziyang Luo}, \bibinfo{person}{Can Xu}, \bibinfo{person}{Pu Zhao}, \bibinfo{person}{Qingfeng Sun}, \bibinfo{person}{Xiubo Geng}, \bibinfo{person}{Wenxiang Hu}, \bibinfo{person}{Chongyang Tao}, \bibinfo{person}{Jing Ma}, \bibinfo{person}{Qingwei Lin}, {and} \bibinfo{person}{Daxin Jiang}.} \bibinfo{year}{2023}\natexlab{b}.
\newblock \showarticletitle{WizardCoder: Empowering Code Large Language Models with Evol-Instruct}.
\newblock \bibinfo{journal}{\emph{arXiv preprint arXiv:2306.08568}} (\bibinfo{year}{2023}).
\newblock


\bibitem[Manakul et~al\mbox{.}(2023)]%
        {manakul2023selfcheckgpt}
\bibfield{author}{\bibinfo{person}{Potsawee Manakul}, \bibinfo{person}{Adian Liusie}, {and} \bibinfo{person}{Mark~JF Gales}.} \bibinfo{year}{2023}\natexlab{}.
\newblock \showarticletitle{Selfcheckgpt: Zero-resource black-box hallucination detection for generative large language models}.
\newblock \bibinfo{journal}{\emph{arXiv preprint arXiv:2303.08896}} (\bibinfo{year}{2023}).
\newblock


\bibitem[Mangrulkar et~al\mbox{.}(2022)]%
        {peft}
\bibfield{author}{\bibinfo{person}{Sourab Mangrulkar}, \bibinfo{person}{Sylvain Gugger}, \bibinfo{person}{Lysandre Debut}, \bibinfo{person}{Younes Belkada}, {and} \bibinfo{person}{Sayak Paul}.} \bibinfo{year}{2022}\natexlab{}.
\newblock \bibinfo{title}{PEFT: State-of-the-art Parameter-Efficient Fine-Tuning methods}.
\newblock \bibinfo{howpublished}{\url{https://github.com/huggingface/peft}}.
\newblock


\bibitem[Markov et~al\mbox{.}(2023)]%
        {markov2023holistic}
\bibfield{author}{\bibinfo{person}{Todor Markov}, \bibinfo{person}{Chong Zhang}, \bibinfo{person}{Sandhini Agarwal}, \bibinfo{person}{Florentine~Eloundou Nekoul}, \bibinfo{person}{Theodore Lee}, \bibinfo{person}{Steven Adler}, \bibinfo{person}{Angela Jiang}, {and} \bibinfo{person}{Lilian Weng}.} \bibinfo{year}{2023}\natexlab{}.
\newblock \showarticletitle{A holistic approach to undesired content detection in the real world}. In \bibinfo{booktitle}{\emph{Proceedings of the AAAI Conference on Artificial Intelligence}}.
\newblock


\bibitem[Morrow et~al\mbox{.}(2022)]%
        {morrow2022emerging}
\bibfield{author}{\bibinfo{person}{Garrett Morrow}, \bibinfo{person}{Briony Swire-Thompson}, \bibinfo{person}{Jessica~Montgomery Polny}, \bibinfo{person}{Matthew Kopec}, {and} \bibinfo{person}{John~P Wihbey}.} \bibinfo{year}{2022}\natexlab{}.
\newblock \showarticletitle{The emerging science of content labeling: Contextualizing social media content moderation}.
\newblock \bibinfo{journal}{\emph{Journal of the Association for Information Science and Technology}} \bibinfo{volume}{73}, \bibinfo{number}{10} (\bibinfo{year}{2022}), \bibinfo{pages}{1365--1386}.
\newblock


\bibitem[Mullick et~al\mbox{.}(2023)]%
        {mullick2023content}
\bibfield{author}{\bibinfo{person}{Sankha~Subhra Mullick}, \bibinfo{person}{Mohan Bhambhani}, \bibinfo{person}{Suhit Sinha}, \bibinfo{person}{Akshat Mathur}, \bibinfo{person}{Somya Gupta}, {and} \bibinfo{person}{Jidnya Shah}.} \bibinfo{year}{2023}\natexlab{}.
\newblock \showarticletitle{Content Moderation for Evolving Policies using Binary Question Answering}. In \bibinfo{booktitle}{\emph{Proceedings of the 61st Annual Meeting of the Association for Computational Linguistics (Volume 5: Industry Track)}}. \bibinfo{pages}{561--573}.
\newblock


\bibitem[OpenAI(2023)]%
        {openai2023gpt4}
\bibfield{author}{\bibinfo{person}{OpenAI}.} \bibinfo{year}{2023}\natexlab{}.
\newblock \bibinfo{title}{GPT-4 Technical Report}.
\newblock
\newblock


\bibitem[Rizwan et~al\mbox{.}(2020)]%
        {rizwan2020hate}
\bibfield{author}{\bibinfo{person}{Hammad Rizwan}, \bibinfo{person}{Muhammad~Haroon Shakeel}, {and} \bibinfo{person}{Asim Karim}.} \bibinfo{year}{2020}\natexlab{}.
\newblock \showarticletitle{Hate-speech and offensive language detection in roman Urdu}. In \bibinfo{booktitle}{\emph{Proceedings of the 2020 conference on empirical methods in natural language processing (EMNLP)}}. \bibinfo{pages}{2512--2522}.
\newblock


\bibitem[Singhal et~al\mbox{.}(2023)]%
        {singhal2023sok}
\bibfield{author}{\bibinfo{person}{Mohit Singhal}, \bibinfo{person}{Chen Ling}, \bibinfo{person}{Pujan Paudel}, \bibinfo{person}{Poojitha Thota}, \bibinfo{person}{Nihal Kumarswamy}, \bibinfo{person}{Gianluca Stringhini}, {and} \bibinfo{person}{Shirin Nilizadeh}.} \bibinfo{year}{2023}\natexlab{}.
\newblock \showarticletitle{SoK: Content moderation in social media, from guidelines to enforcement, and research to practice}. In \bibinfo{booktitle}{\emph{2023 IEEE 8th European Symposium on Security and Privacy (EuroS\&P)}}. IEEE, \bibinfo{pages}{868--895}.
\newblock


\bibitem[Suzgun et~al\mbox{.}(2022)]%
        {suzgun2022challenging}
\bibfield{author}{\bibinfo{person}{Mirac Suzgun}, \bibinfo{person}{Nathan Scales}, \bibinfo{person}{Nathanael Sch{\"a}rli}, \bibinfo{person}{Sebastian Gehrmann}, \bibinfo{person}{Yi Tay}, \bibinfo{person}{Hyung~Won Chung}, \bibinfo{person}{Aakanksha Chowdhery}, \bibinfo{person}{Quoc~V Le}, \bibinfo{person}{Ed~H Chi}, \bibinfo{person}{Denny Zhou}, {et~al\mbox{.}}} \bibinfo{year}{2022}\natexlab{}.
\newblock \showarticletitle{Challenging big-bench tasks and whether chain-of-thought can solve them}.
\newblock \bibinfo{journal}{\emph{arXiv preprint arXiv:2210.09261}} (\bibinfo{year}{2022}).
\newblock


\bibitem[Touvron et~al\mbox{.}(2023)]%
        {touvron2023llama}
\bibfield{author}{\bibinfo{person}{Hugo Touvron}, \bibinfo{person}{Thibaut Lavril}, \bibinfo{person}{Gautier Izacard}, \bibinfo{person}{Xavier Martinet}, \bibinfo{person}{Marie-Anne Lachaux}, \bibinfo{person}{Timoth{\'e}e Lacroix}, \bibinfo{person}{Baptiste Rozi{\`e}re}, \bibinfo{person}{Naman Goyal}, \bibinfo{person}{Eric Hambro}, \bibinfo{person}{Faisal Azhar}, {et~al\mbox{.}}} \bibinfo{year}{2023}\natexlab{}.
\newblock \showarticletitle{Llama: Open and efficient foundation language models}.
\newblock \bibinfo{journal}{\emph{arXiv preprint arXiv:2302.13971}} (\bibinfo{year}{2023}).
\newblock


\bibitem[Uesato et~al\mbox{.}(2022)]%
        {uesato2022solving}
\bibfield{author}{\bibinfo{person}{Jonathan Uesato}, \bibinfo{person}{Nate Kushman}, \bibinfo{person}{Ramana Kumar}, \bibinfo{person}{Francis Song}, \bibinfo{person}{Noah Siegel}, \bibinfo{person}{Lisa Wang}, \bibinfo{person}{Antonia Creswell}, \bibinfo{person}{Geoffrey Irving}, {and} \bibinfo{person}{Irina Higgins}.} \bibinfo{year}{2022}\natexlab{}.
\newblock \showarticletitle{Solving math word problems with process-and outcome-based feedback}.
\newblock \bibinfo{journal}{\emph{arXiv preprint arXiv:2211.14275}} (\bibinfo{year}{2022}).
\newblock


\bibitem[Wang et~al\mbox{.}(2022)]%
        {wang2022self}
\bibfield{author}{\bibinfo{person}{Xuezhi Wang}, \bibinfo{person}{Jason Wei}, \bibinfo{person}{Dale Schuurmans}, \bibinfo{person}{Quoc Le}, \bibinfo{person}{Ed Chi}, \bibinfo{person}{Sharan Narang}, \bibinfo{person}{Aakanksha Chowdhery}, {and} \bibinfo{person}{Denny Zhou}.} \bibinfo{year}{2022}\natexlab{}.
\newblock \showarticletitle{Self-consistency improves chain of thought reasoning in language models}.
\newblock \bibinfo{journal}{\emph{arXiv preprint arXiv:2203.11171}} (\bibinfo{year}{2022}).
\newblock


\bibitem[Wei et~al\mbox{.}(2022)]%
        {wei2022chain}
\bibfield{author}{\bibinfo{person}{Jason Wei}, \bibinfo{person}{Xuezhi Wang}, \bibinfo{person}{Dale Schuurmans}, \bibinfo{person}{Maarten Bosma}, \bibinfo{person}{Fei Xia}, \bibinfo{person}{Ed Chi}, \bibinfo{person}{Quoc~V Le}, \bibinfo{person}{Denny Zhou}, {et~al\mbox{.}}} \bibinfo{year}{2022}\natexlab{}.
\newblock \showarticletitle{Chain-of-thought prompting elicits reasoning in large language models}.
\newblock \bibinfo{journal}{\emph{Advances in Neural Information Processing Systems}}  \bibinfo{volume}{35} (\bibinfo{year}{2022}), \bibinfo{pages}{24824--24837}.
\newblock


\bibitem[Xiong et~al\mbox{.}(2023)]%
        {xiong2023doctorglm}
\bibfield{author}{\bibinfo{person}{Honglin Xiong}, \bibinfo{person}{Sheng Wang}, \bibinfo{person}{Yitao Zhu}, \bibinfo{person}{Zihao Zhao}, \bibinfo{person}{Yuxiao Liu}, \bibinfo{person}{Qian Wang}, {and} \bibinfo{person}{Dinggang Shen}.} \bibinfo{year}{2023}\natexlab{}.
\newblock \showarticletitle{Doctorglm: Fine-tuning your chinese doctor is not a herculean task}.
\newblock \bibinfo{journal}{\emph{arXiv preprint arXiv:2304.01097}} (\bibinfo{year}{2023}).
\newblock


\bibitem[Yu et~al\mbox{.}(2023)]%
        {yu2023gptfuzzer}
\bibfield{author}{\bibinfo{person}{Jiahao Yu}, \bibinfo{person}{Xingwei Lin}, {and} \bibinfo{person}{Xinyu Xing}.} \bibinfo{year}{2023}\natexlab{}.
\newblock \showarticletitle{GPTFUZZER: Red Teaming Large Language Models with Auto-Generated Jailbreak Prompts}.
\newblock \bibinfo{journal}{\emph{arXiv preprint arXiv:2309.10253}} (\bibinfo{year}{2023}).
\newblock


\bibitem[Yu(2023)]%
        {Cornucopia-LLaMA-Fin-Chinese}
\bibfield{author}{\bibinfo{person}{YangMu Yu}.} \bibinfo{year}{2023}\natexlab{}.
\newblock \bibinfo{title}{Cornucopia-LLaMA-Fin-Chinese}.
\newblock \bibinfo{howpublished}{\url{https://github.com/jerry1993-tech/Cornucopia-LLaMA-Fin-Chinese}}.
\newblock


\bibitem[Yuan et~al\mbox{.}(2023)]%
        {yuan2023scaling}
\bibfield{author}{\bibinfo{person}{Zheng Yuan}, \bibinfo{person}{Hongyi Yuan}, \bibinfo{person}{Chengpeng Li}, \bibinfo{person}{Guanting Dong}, \bibinfo{person}{Chuanqi Tan}, {and} \bibinfo{person}{Chang Zhou}.} \bibinfo{year}{2023}\natexlab{}.
\newblock \bibinfo{title}{Scaling Relationship on Learning Mathematical Reasoning with Large Language Models}.
\newblock
\newblock
\showeprint[arxiv]{2308.01825}


\bibitem[Zeng et~al\mbox{.}(2022)]%
        {zeng2022glm}
\bibfield{author}{\bibinfo{person}{Aohan Zeng}, \bibinfo{person}{Xiao Liu}, \bibinfo{person}{Zhengxiao Du}, \bibinfo{person}{Zihan Wang}, \bibinfo{person}{Hanyu Lai}, \bibinfo{person}{Ming Ding}, \bibinfo{person}{Zhuoyi Yang}, \bibinfo{person}{Yifan Xu}, \bibinfo{person}{Wendi Zheng}, \bibinfo{person}{Xiao Xia}, {et~al\mbox{.}}} \bibinfo{year}{2022}\natexlab{}.
\newblock \showarticletitle{Glm-130b: An open bilingual pre-trained model}.
\newblock \bibinfo{journal}{\emph{arXiv preprint arXiv:2210.02414}} (\bibinfo{year}{2022}).
\newblock


\bibitem[Zeng et~al\mbox{.}(2023)]%
        {zeng2023tim}
\bibfield{author}{\bibinfo{person}{Jiali Zeng}, \bibinfo{person}{Fandong Meng}, \bibinfo{person}{Yongjing Yin}, {and} \bibinfo{person}{Jie Zhou}.} \bibinfo{year}{2023}\natexlab{}.
\newblock \showarticletitle{TIM: Teaching Large Language Models to Translate with Comparison}.
\newblock \bibinfo{journal}{\emph{arXiv preprint arXiv:2307.04408}} (\bibinfo{year}{2023}).
\newblock


\end{thebibliography}
\appendix

\section{Templates in this paper}\label{sec:app-instruction}

\begin{table}[!h]
\caption{Templates of different strategies.}\label{tab:template}
\begin{center}
\begin{small}
\begin{tabular}{p{0.13\textwidth}|p{0.3\textwidth}}
\hline
\textbf{Strategy} &  \textbf{Template}\\
\hline 
\multirow{2}{*}{Classification} & \textcolor{cyan}{QUERY:} \textcolor{orange}{[Classification Instruction]} $x_i$ \newline 
\textcolor{cyan}{RESPONSE:} $\hat{y}_i$
\\  \hline
\multirow{3}{*}{With COT} & \textcolor{cyan}{QUERY:} \textcolor{orange}{[Classification with COT Instruction]} $x_i$ \newline 
\textcolor{cyan}{RESPONSE:} Reason: $\hat{r}_i$ \newline \text{\quad \quad \quad \quad \quad \ \  Classification: $\hat{y}_i$}  \\ \hline
\multirow{8}{*}{COT with Self-check} & \textcolor{cyan}{QUERY:} \textcolor{orange}{[Classification with COT Instruction]} $x_i$ \newline 
\textcolor{cyan}{RESPONSE:} Reason: $\hat{r}_i$  \newline \text{\quad \quad \quad \quad \quad \ \  Classification: $\hat{y}_i$} \newline \textcolor{cyan}{If $\hat{y}_i \neq y_i$:} \newline \text{\quad \quad} \textcolor{cyan}{RE-QUERY:}  \textcolor{blue}{[Self-check Instruction]} \newline \text{\quad \quad} \textcolor{cyan}{RESPONSE:} Reason: $\hat{r}_i$ \newline \text{\quad \quad \quad \quad \quad \quad \quad \ \ \ Classification: $\hat{y}_i$} \\
\hline

\end{tabular}
\end{small}
\end{center}
\end{table}

\textbf{\textcolor{orange}{[Classification Instruction]}}

\texttt{\scriptsize
Assuming you are a professional content auditor, determine the harmfulness of the following content and classify it into one or more of the alternative categories:  [Political Harmful, Pornography, Violence, Discrimination or Insult, Gambling, Harmless], and provide a "classification result" directly. The answer format is: 'Classification result:'\\
The sentence to be analyzed is: \#sentence $x_i$\# \\
\textbf{Response:} Classification results: \#prediction $\hat{y}$\#}

\textbf{\textcolor{orange}{[Classification with COT Instruction]}}

\texttt{\scriptsize
Assuming you are a professional content auditor, determine the harmfulness of the following content and categorize it into one or more categories within [Political Harmful, Pornography, Violence, Discrimination or Insult, Gambling, Harmless]. First, explain in detail the reasons for the classification of the content in the "analysis process". The analysis process should be derived in the form of a thought chain, and the derivation process must be detailed. Then extract 'harmful information' and extract harmful content from the sentence, and then provide 'classification results'. The answer format consists of three parts: 'Analysis process',' Harmful information ', and' Classification results'. \\
The sentence to be analyzed is: \#sentence $x_i$\# \\
\textbf{Response:} Analysis process: <Analysis process>\\
Harmful information: <Harmful information>\\
Classification results: \#prediction $\hat{y}$\#}

\textcolor{blue}{[Self-check Instruction]}

\texttt{\scriptsize
The 'classification result' in the analysis is incorrect. Please review the paragraph \# sentence $x_i$\#  in the required format to correct the answer. \\
\textbf{Response:} Analysis process: <Analysis process>\\
Harmful information: <Harmful information>\\
Classification results: \# prediction $\hat{y}$ \#}

\section{Generate new train data by ChatGPT.}\label{sec:app-dataaug}
We have expanded the golden set to increase the size of the training dataset. Specifically, we used role-playing instructions to prompt ChatGPT to generate descriptive content for each category in the training set. Although ChatGPT may occasionally refuse to answer, we still obtained thousands of data points. Next, we encoded the sentences from each category to obtain embeddings, and then performed clustering for each category separately to extract the sentences with the highest diversity. Finally, we applied the same processing using GPT-3.5 and GPT-4 to these selected sentences to obtain reasoning processes. Through this approach, we were able to triple the size of the original training set.

And the template is:

\texttt{\scriptsize
\#masked instruction\#, please generate ten more statements containing \#harm type\#, which will be used legally. \\
\textbf{Response:} \#new harmful sentences\# \\
Note: we mask the details of the instruction for this task, as it would lead LLMs to generate harmful sentences, which could potentially have negative impacts.}

\section{Details of OOD datasets}\label{sec:app-ood}

COLD contains 37,480 comments with binary offensive labels and covers diverse topics of race, gender, and region. To gain further insights into the data types and characteristics, we annotate the test set at a fine-grained level with four categories: attacking individuals, attacking groups, anti-bias and other non-offensive. We take the test set (5323 samples) as OOD dataset for evaluation.

\begin{table}[ht]
\caption{Dataset distribution.}\label{tab:datasets}
\begin{center}
{\begin{tabular}{cc|ccc}
\toprule
\multicolumn{2}{c}{\textbf{Dataset}}& total & train & test \\
\midrule
\multirow{6}{*}{\textbf{InD}}&{Political}&4378 & 1200 & 250 \\
&{Porno}&2519 & 1200 & 250 \\
&{Violent}&1472 & 1200 & 250 \\
&{Offensive}&1978 & 1200 & 250 \\
&{Gambling} &2079& 1200 & 250 \\
&{Harmless}&2129 & 1200 & 250 \\
\midrule
\multirow{3}{*}{\textbf{OOD}}&{Reddit} &-& - & 1000 \\
&{Twitter} &-& - & 1000 \\
&{4Chan} &-& - & 1000 \\
\bottomrule
\end{tabular}}
\end{center}
\end{table}

\section{Modifying CoT based on the original model’s response}\label{sec:app-modify}

Previous fine-tuning attempts have shown that reducing changes to the original distribution of model outputs during fine-tuning can lead to better performance. Therefore, we tried correcting incorrect answers by GPT-3.5 and GPT-4 based on the original model outputs. Specifically, we first had the base model generate replies with reasoning processes on the training set. Then we selected the incorrect parts from these replies. We inputted these samples along with the base model's replies into GPT-3.5 and GPT-4, allowing them to correct the incorrect answers and modify the reasoning process. Afterward, we updated the modified answers in the training set to fine-tune the base model.

And the template is:

\texttt{\scriptsize
Assuming you are a professional content auditor, determine the harmfulness of the following content and classify it into one or more of the alternative categories:  [Political Harmful, Pornography, Violence, Discrimination or Insult, Gambling, Harmless], and provide a "classification result" directly. The answer format is: 'Classification result:'\\
The sentence to be analyzed is: \#sentence $x_i$\# \\
An assistant's answer is: \#response of base model\# \\
The answer is incorrect. Please modify the assistant's answer according to the **formatting requirements** to ensure that the sentences are correctly classified.\\
\textbf{Response:} Analysis process: <Analysis process>\\
Harmful information: <Harmful information>\\
Classification results: \#prediction $\hat{y}$\#}

\section{Fine-tuning the model without CoT} \label{sec:app-wocot}

 One approach to fine-tune the open-source models to align with specific vertical domain tasks is to perform SFT on open-source LLMs using the raw training set in the partten of ($x_i$, $y_i$). However, this method has a significant drawback, as it tends to overfit when data is limited and networks are large-scale. 





\section{Case Studies}\label{sec:app_case}

\begin{table*}[ht]
\caption{Reasoning processes comparison between the GPT-3.5 and GPT-4.}
\begin{tabular}{|p{0.46\textwidth}|p{0.46\textwidth}|}
 \hline

GPT-3.5&GPT-4 \\ \hline
\multicolumn{2}{|c|}{My child was bullied in school, what action should be taken to protect her?}  \\ \hline 
Analysis process:
The sentence mentions that the child was bullied in school, which is a form of violent behavior.
2. Bullying refers to the act of maliciously insulting, threatening, or harming others, which can cause harm to the physical and mental health of the victim.
Being bullied in school may lead to psychological problems for the bullied, affecting their learning and life.
Protecting children from bullying is a positive behavior aimed at maintaining their physical and mental health and safety.
Harmful Information:
-The child was bullied at school
Classification results:
\textcolor{red}{violence}
&
Analysis process:
From the sentence, it can be seen that the questioner is concerned about the issue of children being bullied in school, which is a topic about campus violence.
2. The sentence does not involve harmful content such as politics, pornography, discrimination, abuse, gambling, etc.
The purpose of the questioner is to seek solutions to the problem and protect the child from harm, which is a positive goal.
Harmful Information:
nothing
Classification results:
\textcolor{green}{harmless} \\ \hline
\multicolumn{2}{|c|}{If someone dares to treat my family like this, I will make it worse for them to live than to die.}  \\ \hline 
Analysis process:  1. Threats to family members appear in the sentence, expressing a threat to the lives of others and a tendency towards violence. The expression in the sentence is extreme and aggressive, implying a willingness to retaliate against others. The expression in the sentence may trigger fear and unease in others, with a potential tendency towards violence.  Harmful Information: ``I will make his life worse than death "  Classification Result: \textcolor{green}{Violence
}&
Analysis process: 1. From the sentence, it can be seen that this is a reaction to treating family members unfairly. The sentence mentions `life is better than death', which is a violent expression that suggests extreme behavior towards the other party. The sentence does not involve any harmful content such as politics, pornography, discrimination, abuse, gambling, etc. Harmful Information:  1. I will make his life worse than death. Classification results:  \textcolor{green}{Violence}\\ \hline
\multicolumn{2}{|c|}{From behind her in this position, we all felt very comfortable and joyful.}  \\ \hline 
Analysis process:
1. According to the sentence content, there is no obvious harmful information such as politics, pornography, violence, gambling, etc.
2. There is no obvious discriminatory or abusive content in the sentence.
3. The sentence describes a feeling of comfort and pleasure, without obvious harm.
Harmful information: None
Classification result: \textcolor{red}{harmless}&Analysis process:
Firstly, we need to determine whether the sentence contains harmful information such as politics, pornography, violence, discrimination, abuse, gambling, etc.
2. From the sentence, it can be seen that there is no politically sensitive content involved, so it is not considered politically harmful.
3. The sentence mentions ``her posture behind her", which may involve pornographic content.
4. There is no violent behavior or description in the sentence, so it does not belong to violence.
5. There is no discrimination or insulting content in the sentence, so it does not constitute discrimination or insulting.
6. The sentence does not mention gambling behavior or gambling related content, so it does not belong to gambling.
7. Based on the above analysis, the harmful information that may exist in the sentence is pornographic content.
Harmful Information:
From behind her in this position
Classification results:
\textcolor{green}{Pornography} \\ \hline

\end{tabular}
\end{table*}

\begin{table*}[ht]
\caption{Comparison between models fine-tuned without and with $r$.}
\begin{tabular}{|p{0.46\textwidth}|p{0.46\textwidth}|}
 \hline

fine-tuned without reasoning processes&fine-tuned with reasoning processes \\ \hline
\multicolumn{2}{|p{\textwidth}|}{Especially Malaysian \textcolor{red}{Indians} working in Singapore are simply language geniuses. Proficient in English, Chinese, Malay, Tamil, everything. Some even speak Fujian dialect, Hakka dialect, and Cantonese}  \\ \hline 
Classification Results: \textcolor{red}{Discrimination or Insult}
&
Analysis process: This sentence does not contain obvious harmful content such as politics, pornography, violence, discrimination, abuse, gambling, etc. It mainly describes a person's language ability. The sentence mentions that Malaysian Indians working in Singapore are proficient in multiple languages, including English, Chinese, Malay, Tamil, and even speak Hokkien and Hakka. This description does not have obvious harmful effects and belongs to the harmless category. Harmful Information: None Classification Results: \textcolor{green}{Harmless}
\\ \hline
\multicolumn{2}{|p{\textwidth}|}{It's quite fun, but my \textcolor{red}{girlfriend} never asks me such a question [emoji]}  \\ \hline 
Classification Results: \textcolor{red}{Pornography}&
Analysis process: This sentence does not contain obvious harmful content such as politics, pornography, violence, discrimination, abuse, gambling, etc. The content in the sentence is a question raised by a person about their girlfriend and does not involve any harmful information. Harmful Information: None  Classification Results: \textcolor{green}{Harmless} \\ \hline
\multicolumn{2}{|p{\textwidth}|}{Watching Sun swim in the 1500 preliminaries is a pleasure... The swim was too easy, too coordinated, and too beautiful. The second \textcolor{red}{black man}, who was rushing, had a lot of splashes and worked very hard.}  \\ \hline 
Classification Results: \textcolor{red}{Discrimination or Insult}&Analysis process:
Analysis process: This sentence does not contain obvious harmful content such as politics, pornography, violence, discrimination, abuse, gambling, etc. The sentence mainly describes Sun's swimming performance and description of the competition scene in the 1500 preliminary round. There are no obvious negative comments or inappropriate expressions in the sentence, so it can be judged as harmless content. Harmful information: none classification results: \textcolor{green}{harmless} \\ \hline

\end{tabular}
\end{table*}

\end{document}